\RequirePackage{fix-cm}
\documentclass[twocolumn]{svjour3}          
\smartqed  
\usepackage{times}
\usepackage{epsfig}
\usepackage{graphicx}
\usepackage{amsmath}
\usepackage{amssymb}
\usepackage{adjustbox} 
\usepackage[]{algorithm2e}
\usepackage{etoolbox}
\usepackage{multirow}
\usepackage{booktabs,siunitx} 
\usepackage{color}
\usepackage{subfig}
\usepackage{comment}
\usepackage{gensymb}
\usepackage{pifont}
\usepackage{mdframed}
\usepackage{makecell} 
\setcellgapes{5pt}
\DeclareMathOperator*{\argmax}{argmax}

\makeatletter
\newcommand\footnoteref[1]{\protected@xdef\@thefnmark{\ref{#1}}\@footnotemark}

\makeatother

\usepackage[pagebackref=true,breaklinks=true,colorlinks,bookmarks=true]{hyperref}

\journalname{International Journal of Computer Vision}
\begin{document}

\title{Talk2Nav: Long-Range Vision-and-Language Navigation with Dual Attention and Spatial Memory  
}


\author{Arun Balajee Vasudevan \and
        Dengxin Dai  \and  Luc Van Gool 
}


\institute{Arun Balajee Vasudevan         \and
        Dengxin Dai  \and  Luc Van Gool \at
              ETH Zurich, Switzerland\\
              \email{\{arunv, dai, vangool\}@vision.ee.ethz.ch}           
           \and
           Luc Van Gool \at
              K.U Leuven, Belgium
}

\date{Received: 13 August 2019 / Accepted: 19 August 2020}

\maketitle

\begin{abstract}
The role of robots in society keeps expanding, bringing with it the necessity of interacting and communicating with humans. 
In order to keep such interaction intuitive, we provide automatic wayfinding based on verbal navigational instructions. Our first contribution is the creation of a large-scale dataset with verbal navigation instructions. To this end, we have developed an interactive visual navigation environment based on Google Street View; we further design an annotation method to highlight mined anchor landmarks and local directions between them in order to help annotators formulate typical, human references to those. The annotation task was crowdsourced on the AMT platform, to construct a new  Talk2Nav dataset with $10,714$ routes. Our second contribution is a new learning method. Inspired by spatial cognition research on the mental conceptualization of navigational instructions,
we introduce a soft dual attention mechanism defined over the segmented language instructions to jointly extract two partial instructions -- one for matching the next upcoming visual landmark and the other for matching the local directions to the next landmark.
On the similar lines, we also introduce spatial memory scheme to encode the local directional transitions.
Our work takes advantage of the advance in two lines of research: mental formalization of verbal navigational instructions and training neural network agents for automatic way finding. 
Extensive experiments show that our method significantly outperforms previous navigation methods. 
For demo video, dataset and code, please refer to our \href{https://www.trace.ethz.ch/publications/2019/talk2nav/index.html}{project page}.
\end{abstract}

\section{Introduction}
\label{sec:intro}
Consider that you are traveling as a tourist in a new city and are looking for a destination that you would like to visit. You ask the locals and get a directional description ``\emph{go ahead for about 200 meters until you hit a small intersection, then turn left and continue along the street before you see a yellow building on your right}''. People give indications that are not purely directional, let alone metric. They mix in referrals to landmarks that you will find along your route. This may seem like a trivial ability, as humans do this routinely. Yet, this is a complex cognitive task that relies on the development of an internal, spatial representation that includes visual landmarks (e.g. ``the yellow building'') and possible, local directions (e.g. ``going forward for about 200 meters''). Such representation can support a continuous self-localization as well as conveying a sense of direction towards the goal. 

Just as a human can navigate in an environment when provided with navigational instructions, our aim is to teach an agent to perform the same task to make the human-robot interaction more intuitive.
The problem is tackled recently as a Vision-and-Language Navigation (VLN) problem~\cite{anderson2018vision}. Although important progress has been made, e.g. in constructing good datasets~\cite{anderson2018vision,chen2018touchdown} and proposing effective learning methods ~\cite{anderson2018vision,speaker:follower:nips18,look:before:leap:eccv18,ma2019regretful}, this stream of work mainly focuses on synthetic worlds~\cite{Hermann2017GroundedLL,embodiedqa:cvpr18,chen2011learning} or indoor room-to-room navigation~\cite{anderson2018vision,speaker:follower:nips18,look:before:leap:eccv18}. Synthetic environments limit the complexity of the visual scenes while the room-to-room navigation comes with the kind of challenges different from those of outdoors.

\begin{figure*}[t]
\includegraphics[width=1.0\linewidth,trim={.00\textwidth} {.08\textwidth} {.00\textwidth} {.00\textwidth},clip]
    {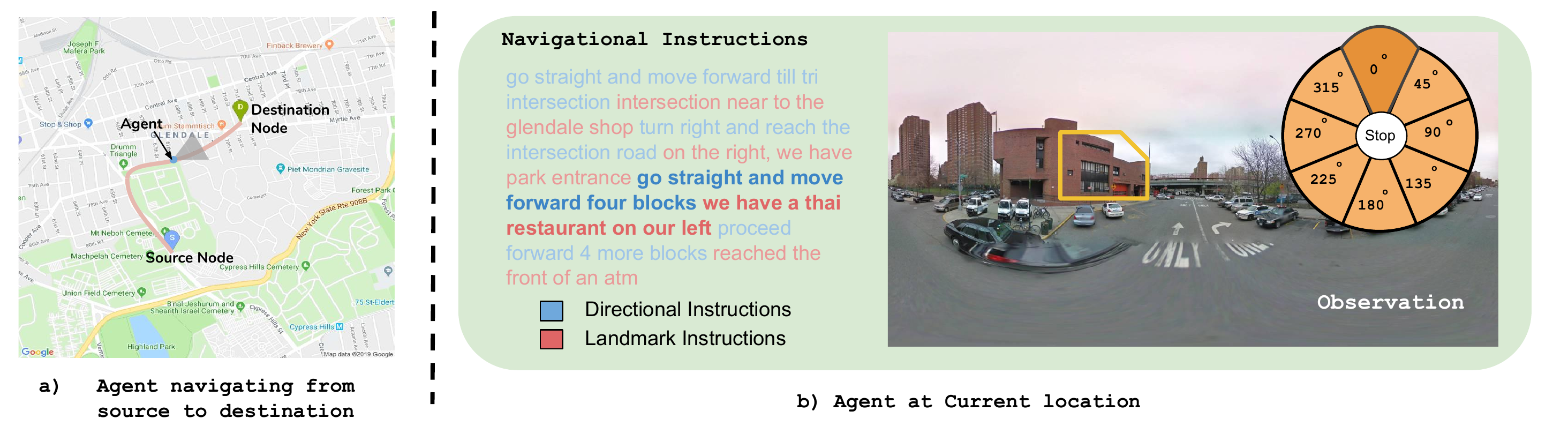} \\
    \text{ \: (a) Top view of the navigation route \qquad \qquad \qquad \qquad  \qquad \qquad (b)  Street view of the route and the agent's status} 
    \caption{ (a) An illustration of an agent finding its way from a source point to a destination. (b) Status of the agent at the current location including the segmented navigational instructions with \textcolor{red}{red} indicating visual landmark descriptions and \textcolor{blue}{blue} the local directional instructions, and the agent's panoramic view and a pie chart depicting the action space. The bold text represents currently attended description.  The predicted action at the current location is highlighted. The landmark that the agent is looking for at the moment is indicated by a yellow box. } 
\label{fig:teaser}
\end{figure*}


The first challenge to learning a long-range wayfinding model lies in the creation of large-scale datasets. In order to be fully effective, the annotators providing the navigation instructions ought to know the environment like locals would. Training annotators to reach the same level of understanding for a large number of unknown environments is inefficient -- to create one verbal navigation instruction, an annotator needs to search through hundreds of street-view images, remember their spatial arrangement, and summarize them into a sequence of route instructions. This straightforward annotation approach would be very time-consuming and error-prone. Because of this challenge, the state-of-the-art work uses synthetic directional instructions~\cite{learning:to:follow:19} or works mostly on indoor room-to-room navigation. For indoor room-to-room navigation, this challenge is less severe, due to two reasons: 1) the paths in indoor navigation are shorter; and 2) indoor environments have a higher density of familiar `landmarks'. This makes self-localization, route remembering and describing easier. 

To our knowledge, there is only one other work by Chen \textit{et al.}~\cite{chen2018touchdown} on natural language based outdoor navigation, which also proposes an outdoor VLN dataset. Although they have designed a great method for  data annotation through gaming -- to find a hidden object at the goal position, the method has difficulty to be applied to longer routes (see discussion in Section~\ref{sec:anno:dataset:statstics}). Again, it is very hard for the annotators to remember, summarize and describe long routes in an unknown environment.    


To address this challenge, we draw on the studies in cognition and psychology on human visual navigation  which state the significance of using visual landmarks in route descriptions~\cite{holscher2011would,ishikawa2012landmark,tom2004language}. It has been found that route descriptions consist of descriptions for visual landmarks and local directional instructions between consecutive landmarks~\cite{landmark-based:navigation:07,when:why:visual:landmarks:01}. Similar techniques -- a combination of visual landmarks, as rendered icons, and highlighted routes between consecutive landmarks -- are constantly used for making efficient maps~\cite{pictorial:verbal:tools:99,weissenberg2014navigation,grabler2008automatic}. We divide the task of generating the description for a whole route into a number of  sub-tasks consist of generating descriptions for visual landmarks and generating local directional instructions between consecutive landmarks. This way, the annotation tasks become simpler.      


We develop an interactive visual navigation environment based on Google Street View, and more importantly design a novel annotation method which highlights selected landmarks and the spatial transitions in between. This enhanced annotation method makes it feasible to crowdsource this complicated annotation task. By hosting the tasks on the Amazon Mechanical Turk (AMT) platform, this work has constructed a new dataset \emph{Talk2Nav} with $10,714$ long-range routes within New York City (NYC). 

The second challenge lies in training a long-range wayfinding agent. This learning task requires accurate visual attention and language attention, accurate self-localization and a good sense of direction towards the goal. 
Inspired by the research on mental conceptualization of navigational instructions in spatial cognition~\cite{pictorial:verbal:tools:99,when:why:visual:landmarks:01,structural:salience:landmarks:route:directions:05}, we introduce a soft attention mechanism defined over the segmented language instructions to jointly extract two partial instructions -- one for matching the next coming visual landmark and the other for matching the spatial transition to the next landmark. Furthermore, the spatial transitions of the agent are encoded by an explicit memory framework which can be read from and written to as the agent navigates. One example of the outdoor VLN task can be found in Figure~\ref{fig:teaser}. Our work connects two lines of research that have been less explored together so far: mental formalization of verbal navigational instructions~\cite{pictorial:verbal:tools:99,when:why:visual:landmarks:01,structural:salience:landmarks:route:directions:05} and training neural network agent for automatic wayfinding~\cite{anderson2018vision,learning:to:follow:19}. 

Extensive experiments show that our method outperforms previous methods by a large margin. We also show the contributions of the sub-components of our method, accompanied with their detailed ablation studies. The collected dataset will be made publicly available.

\section{Related Works}
\label{sec:related}
\textbf{Vision \& Language}. 
Research at the intersection of language and vision has been conducted extensively in the last few years. The main topics include image captioning~\cite{karpathy2015deep,show:attend:tell}, visual question answering (VQA)~\cite{agrawal2017vqa,andreas2016learning}, object referring expressions~\cite{deruyttere2019talk2car,hendricks2017localizing,ORVideoGaze,vasudevan2018object}, and grounded language learning~\cite{Hermann2017GroundedLL,gounded:language:agent:17}. Although the goals are different from ours, some of the fundamental techniques are shared. For example, it is a common practice to represent visual data with CNNs pre-trained for image recognition and to represent textual data with word embeddings pre-trained on large text corpora. The main difference is that the perceptual input to the system is static while ours is active, i.e. the system’s behavior changes the perceived input. 

\noindent
\textbf{Vision Based Navigation}.
Navigation based on vision and reinforcement learning (RL) has become a very interesting research topic recently.  
The technique has proven quite successful in simulated environments~\cite{Learning:to:navigate:iclr17,zhu2017icra} and is being extended to more sophisticated real environments~\cite{navigation:wo:map:18}. 
There has been active research on navigation-related tasks, such as localizing from only an image~\cite{plaNet:eccv16}, finding the direction to the closest
McDonald’s, using Google Street View Images~\cite{khosla2014looking,brahmbhatt2017deepnav}, goal based visual navigation~\cite{gupta2019cognitive} and others. Gupta \textit{et al.}~\cite{gupta2019cognitive} uses a differentiable mapper which writes into
a latent spatial memory corresponding to an egocentric map of the environment and a differentiable planner which uses this memory and the given goal to give navigational actions to navigate in novel environments. There are few other recent works on vision based navigation ~\cite{Thoma_2019_CVPR,Wortsman_2019_CVPR}. Thoma \textit{et al.}~\cite{Thoma_2019_CVPR} formulates compact map construction and accurate self localization for image-based navigation by a careful selection of suitable visual landmarks. Recently, Wortsman \emph{et al.}~\cite{Wortsman_2019_CVPR} proposes a meta-reinforcement learning approach for visual navigation, where the agent learns to adapt in unseen environments in a self-supervised manner. There is another stream of work called end-to-end driving, aiming to learn vehicle navigation directly from video inputs~\cite{drive:surroundview:route:planner,human-like:driving:19}.   


\noindent
\textbf{Vision-and-Language Navigation}.
The task is to navigate an agent in an environment to a particular destination based on language instructions. The following are some recent works in Vision-and-Language Navigation (VLN)~\cite{anderson2018vision,look:before:leap:eccv18,speaker:follower:nips18,wang2019reinforced,Nguyen_2019_CVPR,ma2019regretful,ke2019tactical} task. 
The general goal of these works are similar to ours -- to navigate from a starting point to a destination in a visual environment with language directional descriptions. Anderson \textit{et al.}~\cite{anderson2018vision} created the R2R dataset for indoor room-to-room navigation and proposed a learning method based on sequence-to-sequence neural networks. Subsequent methods \cite{look:before:leap:eccv18,wang2019reinforced} apply reinforcement learning and cross modal matching techniques on the same dataset. The same task was tackled by Fried \textit{et al.}~\cite{speaker:follower:nips18} using speaker-follower technique to generate synthetic instructions for data augmentation and pragmatic inference. While sharing similarity, our work differs significantly from them. The environment domain is different; as discussed in Section~\ref{sec:intro}, long-range navigation raises more challenges both in data annotation and model training. There are concurrent works aiming at extending Vision-and-Language Navigation to city environment ~\cite{learning:to:follow:19,chen2018touchdown,Kim_2019_CVPR}. The difference to Hermann \textit{et al.}~\cite{learning:to:follow:19} lies in that our method works with real navigational instructions, instead of the synthetic ones summarized by Google Maps. This difference leads to different tasks and in turn to different solutions. Kim \textit{et al.}~\cite{Kim_2019_CVPR} proposes end-to-end driving model that takes natural language advice to predict control commands to navigate in city environment.  Chen \textit{et al.}~\cite{chen2018touchdown} proposes outdoor VLN dataset similar to ours, where real instructions are created from Google Street View\footnote{https://developers.google.com/streetview/} images. However, our method differs in the way that it decomposes the navigational instructions to make the annotation easier. It draws inspiration from spatial cognition field to specifically promote annotators' memory and thinking, making the task less energy-consuming and less error-prone. More details can be found in Section~\ref{sec:dataset}.        


\noindent
\textbf{Attention for Language Modeling}.
Attention mechanism has been used widely for language~\cite{mansimov2015generating} and visual modeling~\cite{show:attend:tell,wang2017residual,anderson2018bottom}.
Language attention mechanism has been shown to produce state-of-the-art results in machine translation~\cite{bahdanau2014neural} and other natural language processing tasks like VQA~\cite{hudson2018compositional,yang2016stacked,hu2018explainable}, image captioning~\cite{aneja2018convolutional}, grounding referential expressions~\cite{hu2018explainable,hu2019language} and others. Attention mechanism is one of the main component for the top-performing algorithms such as Transformer~\cite{vaswani2017attention} and BERT~\cite{devlin2018bert} in NLP tasks. The MAC by Hudson \textit{et al.}~\cite{hudson2018compositional} has a control unit which performs weighted average of the question words based on a soft attention for VQA task. Hu \textit{et al.}~\cite{hu2018explainable} also proposed a similar language attention mechanism but they decomposed the reasoning into sub-tasks/modules and predicted modular weights from the input text.
In our model, we adapt the soft attention proposed by Kumar \textit{et al.}~\cite{kumar2018visual} by applying the soft attention over segmented language instructions to put the attention over two different sub-instructions: a) for landmarks and b) for local directions. The attention mechanism is named dual attention. 

\noindent
\textbf{Memory}.
There are generally two kinds of memory used in the literature: a) implicit memory and b) explicit memory.  Implicit memory learns to memorize knowledge in the hidden state vectors via back-propagation of errors. Typical examples include RNNs~\cite{karpathy2015deep} and LSTMs~\cite{donahue2015long}. Explicit memory, however, features explicit read and write modules with an attention menchanism. Notable examples include Neural Turing Machines~\cite{graves2014neural} and Differentiable Neural Computers (DNCs)~\cite{graves2016hybrid}. In our work, we use external explicit memory in the form of a memory image which can be read from and written to by its read and write modules. 
Training a soft attention mechanism over language segments coupled with an explicit memory scheme makes our method more suitable for long-range navigation where the reward signals are sparse.

\section{Talk2Nav Dataset}
\label{sec:dataset}
The target is to navigate using the language descriptions in real outdoor environment. Recently, a few datasets on language based visual navigation task have been released both for indoor~\cite{anderson2018vision} and outdoor~\cite{boularias2015grounding,chen2018touchdown} environments. Existing datasets typically annotate one overall language description for the entire path/route. This poses challenges for annotating longer routes as stated in Section~\ref{sec:intro}. Furthermore, these annotations lack the correspondence between language descriptions and sub-units of a route. This in turn poses challenges in learning long-range vision-and-language navigation (VLN). To address the issue, this work proposes a new annotation method and uses it to create a new dataset Talk2Nav. 

\begin{figure}
\includegraphics[width=1.0\linewidth,trim={.00\textwidth} {.00\textwidth} {.00\textwidth} {.00\textwidth},clip]{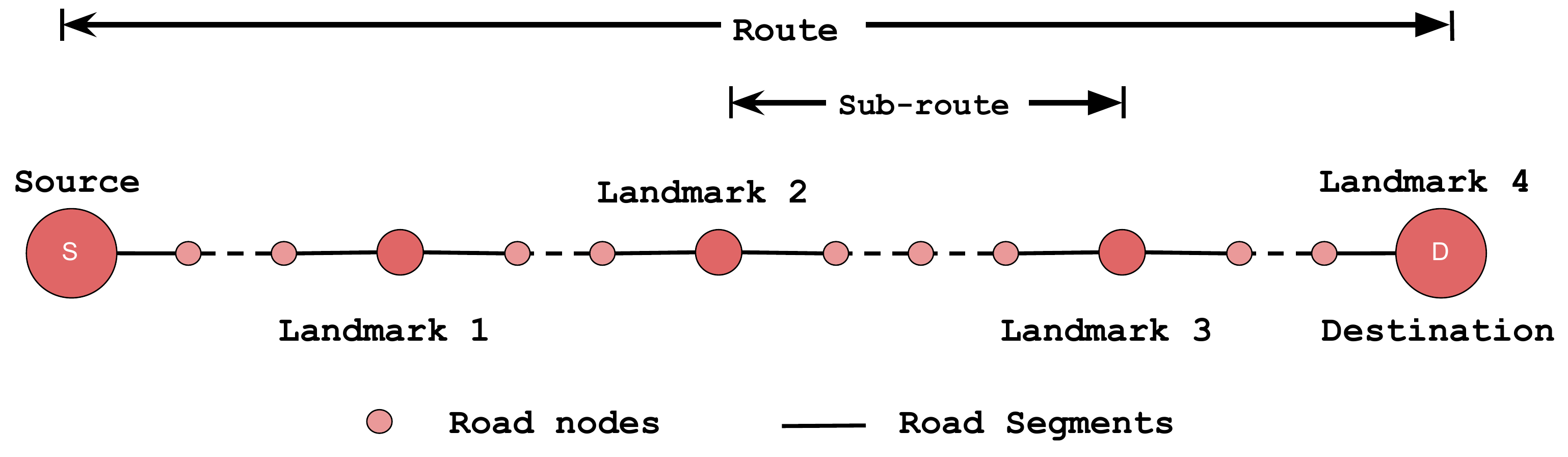}
\caption{An illustrative route from a source node to a destination node with all landmarks labelled along the route. In our case, the destination is landmark 4. Sub-routes are defined as the routes between two consecutive landmarks.}
\label{fig:route_description}
\end{figure}

Talk2Nav contains navigation routes at city levels. A navigational city graph is created with nodes as locations in the city and connecting edges as the road segments between the nodes, which is similar to Mirowski \textit{et al.}~\cite{navigation:wo:map:18}. A route from a source node to a destination node is composed of densely sampled atomic unit nodes. Each node contains a) a street-view panoramic image, b) the GPS coordinates of that location, c) the bearing angles to connecting edges (roads). Furthermore, we enrich the routes with intermediary visual landmarks, language descriptions for these visual landmarks and the local directional instructions for the sub-routes connecting the landmarks. An illustration of a route is shown in  Figure~\ref{fig:route_description}.


\subsection{Data Collection}
\label{dataset:collection}
To retrieve city route data, we used OpenStreetMap to obtain the metadata information of locations and Google's APIs for maps and street-view images.

\bigskip
\noindent
\textbf{Path Generation}. The path from a source to a destination is sampled from a city graph. To that aim, we used OpenStreetMap (OSM) which provides latitudes, longitudes and bearing angles of all the locations (waypoints) within a predefined region in the map. 
We use Overpass API to extract all the locations from OSM, given a region.
A city graph is defined by taking the locations of the atomic units as the nodes, and the directional connections between neighbourhood locations as the edges. 
The K-means clustering algorithm ($k=5$) is applied to the GPS coordinates of all nodes. We then randomly picked up two nodes from different clusters to ensure that the source node and the destination node are not too close. We then use opensourced OpenStreet Routing Machine to extract a route plan from every source node to the destination node, which are sampled from the set of all compiled nodes.
The $A*$ search algorithm is then used to generate a route by finding the shortest traversal path from the source to the destination in the city graph. 

\begin{figure*}[t] \vspace{0mm}
\includegraphics[width=1\linewidth,trim={.00\textwidth} {.00\textwidth} {0.00\textwidth} {.00\textwidth},clip]{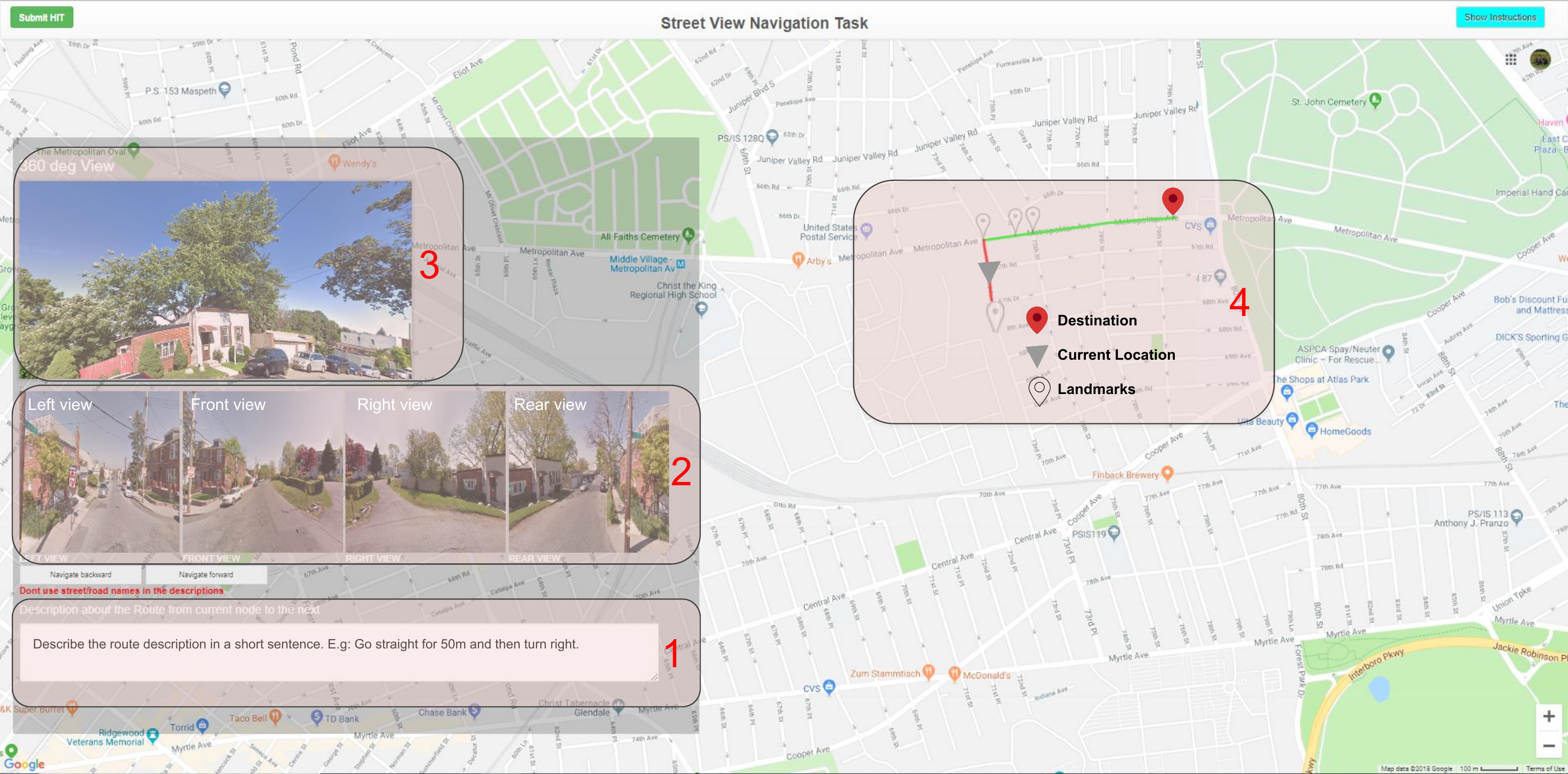} 
\caption{The annotation interface used in Mechanical Turk. Box 1 represents the text box where the annotators write the descriptions. Box 2 denotes the perspective projected images of left, front, right and rear view at the current location. Box 3 shows the Street View environment where the annotator can drag and rotate to get 360\degree view. Box 4 represents the path to-be-annotated with the markers for the landmarks. Red line in Box 4 denotes the current subroute. We can navigate forward and backward along the marked route (lines drawn in red and green), perceiving the Street View simultaneously on the left. Boxes are provided for describing the marked route for landmarks and directions between them. Zoom In for a better view.}
\label{fig:interface}
\end{figure*}

\bigskip
\noindent
\textbf{Street View Images}.
Once the routes are generated using OSM, we densely interpolated the latitudes and longitudes of the locations along the routes to extract more road locations. We then use Google Street View APIs to get the closest StreetView panorama and omit the locations which do not have a streetview image. 
We collect the $360\degree$ street-view images along with their heading angles by using Google Street View APIs and the metadata\footnote{http://maps.google.com/cbk?output=xml\&ll=40.735357,-73.918551\&dm=1}.  The API allows for downloading tiles of a street-view panoramic image which are then stitched together to get the equirectangular projection image. We use the heading angle to re-render the street-view panorama image such that the image is centre-aligned to the heading directions of the route.

We use OSM and opensourced OSM related APIs for the initial phase of extraction of road nodes and route plans. However later, we move to Google Street View for mining streetview images and their corresponding metadata. We noted that it was not straightforward to mine the locations/road nodes from Google Street View.

\subsection{Directional Instruction Annotation}
The main challenge of data annotation for automatic language-based wayfinding lies in the fact that the annotators need to play the role of an instructor as the local people do to tourists. This is especially challenging when the annotators do not know the environment well.  The number of street-view images for a new environment is tremendous and searching
through them can be costly, not to mention remembering and summarizing them to verbal directional instructions. 

Inspired by the large body of work in cognitive science on how people mentally conceptualize route information and convey routes~\cite{wayfinding:choremes:05,when:why:visual:landmarks:01,pictorial:verbal:tools:99}, our annotation method is designed to specifically promote memory or thinking of the annotators. For the route directions, people usually refer to visual landmarks~\cite{holscher2011would,when:why:visual:landmarks:01,tom2003referring} along with a local directional instruction~\cite{pictorial:verbal:tools:99,vogel2010learning}. 
Visualizing the route with highlighted salient landmarks and local directional transitions compensates for limited familiarity or understanding of the environment. 

\subsubsection{Landmark Mining}
\label{subsec:ranking}
The ultimate goal of this work is to make human-to-robot interaction more intuitive, thus the instructions need to be similar to those used by daily human communication.   
Simple methods of mining visual landmarks based on some CNN features may lead to images which are hard for human to distinguish and describe. Hence, this may lead to low-quality or unnatural instructions.

We frame the landmark mining task as a summarization problem using sub-modular optimization to create summaries that takes into account multiple objectives. In this work, three criteria are considered: 1) the selected images are encouraged to spread out along the route to support continuous localization and guidance; 2) images close to road intersections and the approaching side of the intersections are preferred for better guidance through the intersections; and 3) images which are easy to be described, remembered and identified are preferred for effective communication.

\begin{figure*}[t]
\includegraphics[width=1.0\linewidth,trim={.00\textwidth} {.08\textwidth} {0.0\textwidth} {.08\textwidth},clip]{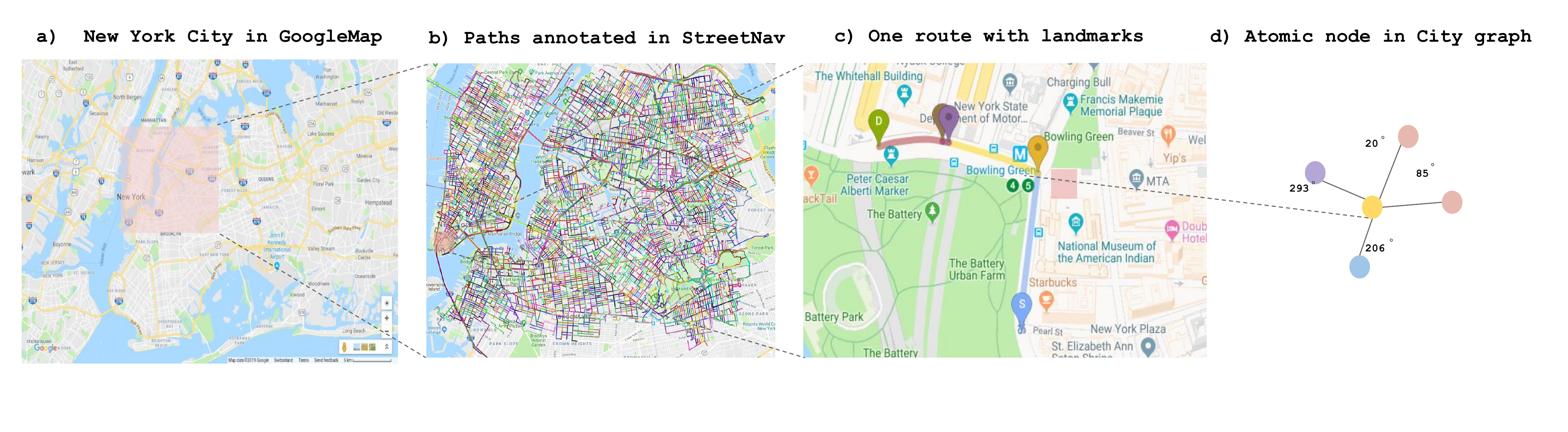} \\
\text{\quad  (a) the annotated region in NYC \quad  \quad  \quad  (b) the created city graph  \quad \quad (c) a route with landmarks and sub-routes \quad   (d) the structure of a node}
\caption{An illustration of our Talk2Nav dataset at multiple granularity levels. }
\label{fig:dataset_structure}
\end{figure*}


Given the set of all images $\mathcal{I}$ along a path $P$, the problem is formulated as a subset selection problem that maximizes a linear combination of the three submodular objectives: 
\begin{equation}
\label{eq:eqn1}
 \mathcal{L} =  \argmax_{\mathcal{L}^\prime \subseteq \wp{(\mathcal{I}})} \sum_{i=1}^{3} w_{i}f_{i}(\mathcal{L}^\prime, P), \quad  s.t. |\mathcal{L}^\prime|=l
\end{equation}
where $\wp{(\mathcal{I}})$ is the powerset of $\mathcal{I}$,  $\mathcal{L}^\prime$ is the set of all possible solutions for the size of $l$, $w_{i}$ are non-negative weights, and $f_{i}$ are the sub-modular objective functions. More specifically, $f_1$ is the minimum travel distance between any of the two consecutive selected images along the route $P$; $f_2=1/(d+\sigma)$ with $d$ the distance to the closest approaching intersection and $\sigma$ is set to $15$ meters to avoid having an infinitely large value for intersection nodes; $f_3$ is a learned ranking function which signals the easiness of describing and remembering the selected images. The weights $w_{i}$ in Equation~\ref{eq:eqn1} are set empirically: $w_1=1$, $w_2=1$ and $w_3=3$. $l$ is set to $3$ in this work as we have the source and the destination nodes fixed and we choose three intermediate landmarks as shown in Figure~\ref{fig:route_description}. Routes of this length with $3$ intermediate landmarks are already long enough for the current navigation algorithms. 
The function $f_3$ is a learned ranking model, which is presented in the next paragraph. 




\smallskip
\noindent

\bigskip
\noindent
\textbf{Ranking model}. In order to train the ranking model for images of being `visual landmarks' that are easier to describe and remember, we compile images from three cities: New York City (NYC), San Francisco (SFO) and London covering different scenes such as high buildings, open fields, downtown areas, etc. We select 20,000 pairs of images from the compiled set. A pairwise comparison is performed over 20,000 pairs to choose one over the other for the preferred landmark. We crowd-sourced the annotation with the following criteria: a)  Describability -- how easy to describe it by the annotator, b) Memorability -- how easy to remember it by the agent (e.g. a traveler such as a tourist), and c) Recognizability -- how easy to recognize it by the agent. Again, our ultimate goal is to make human-to-robot interaction more intuitive, so these criteria are inspired by how human select landmarks when formulating navigational descriptions in our daily life.  

We learn the ranking model with a Siamese network~\cite{chopra2005learning} by following~\cite{gygli2016video2gif}. The model takes a pair of images and scores the selected image more than the other one. We use the Huber rank loss as the cost function. In the inference stage, the model ($f_{3}$ in Equation~\ref{eq:eqn1}) outputs the averaged score for all selected images signalling their suitability as visual landmarks for the VLN task. 

\subsubsection{Annotation and Dataset Statistics}
\label{sec:anno:dataset:statstics}

In the literature, a few datasets have been created for similar tasks~\cite{anderson2018vision,chen2018touchdown,de2018talk,zhangBRN}. 
For instance, Anderson \textit{et al.}~\cite{anderson2018vision} annotated the language description for the route by asking the user to navigate the entire path in egocentric perspective. Incorporation of overhead map of navigated route as an aid for describing the route can be seen in~\cite{chen2018touchdown,de2018talk,zhangBRN}. 

In the initial stages of our annotation process, we followed the above works. We allowed the annotators to navigate the entire route and describe a single instruction for the complete navigational route. We observed that a) the descriptions are mostly about the ending part of the routes indicating that the annotators forget the earlier stages of the routes; b) the annotators took a lot of time to annotate as they have to move back and forth multiple times, c) the annotation errors are very high. 
These observations confirmed our conjecture that it is very challenging to create high-quality annotations for large-scale, long-range VLN tasks. For each annotation, the annotator needs to know the environment as well as the locals would in order to be accurate. This is time-consuming and error-prone. To address this issue, we simplify the annotation task.

In our route description annotation process, the mined visual landmarks are provided, along with an overhead topological map and the street-view interface. 
We crowd-source this annotation task on Amazon Mechanical Turk. In the interface, a pre-defined route on the GoogleMap is shown to the annotator. An example of the interface is shown in Figure~\ref{fig:interface}. The green line denotes the complete route while the red line shows the current road segment. We use \textit{Move forward} and \textit{Move backward} button to navigate from the source node to the destination node. The annotator is instructed to watch the 360\degree Street-view images on the left. Here, we have customized the Google Street View interface to allow the annotator to navigate along the street-view images simultaneously as they move forward/backward in the overhead map. The street view is aligned to the direction of the navigation such that \emph{forward} is always aligned with the moving direction. To minimize the effort of understanding the street-view scenes, we also provide four perspective images for the left-view, the front-view, the right-view, and the rare-view. 

The annotator can navigate through the complete route comprising of $m$ landmark nodes and $m$ intermediate sub-routes and is asked to provide descriptions for all landmarks and descriptions for all sub-routes. 
In this work, we use $m=4$, which means that we collect $4$ landmark descriptions and $4$ local directional descriptions for each route as shown in Figure~\ref{fig:route_description}. We then append the landmark descriptions and the directional descriptions for the sub-routes one after the other to yield the complete route description. 
The whole route description is generated in a quite formulaic way which may lead to not very fluent descriptions. The annotation method offers an efficient and reliable solution at a modest price of language fluency. 




\begin{figure}[!tb]
\centering
\includegraphics[width=0.495\linewidth,trim={.02\textwidth} {.00\textwidth} {0.05\textwidth} {.00\textwidth},clip]{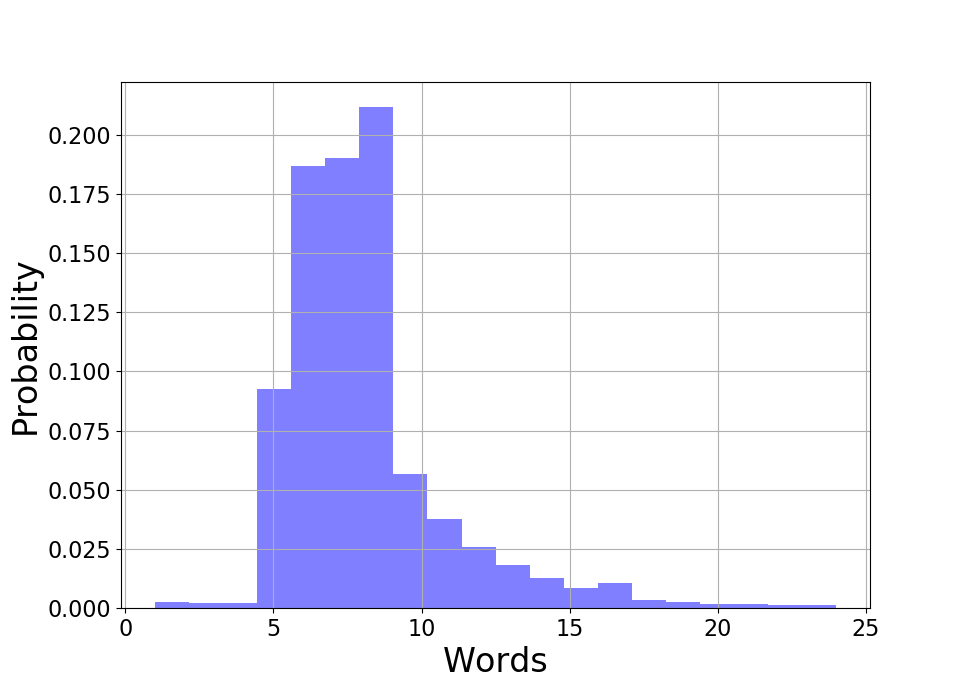}
\includegraphics[width=0.495\linewidth,trim={.03\textwidth} {.00\textwidth} {0.05\textwidth} {.00\textwidth},clip]{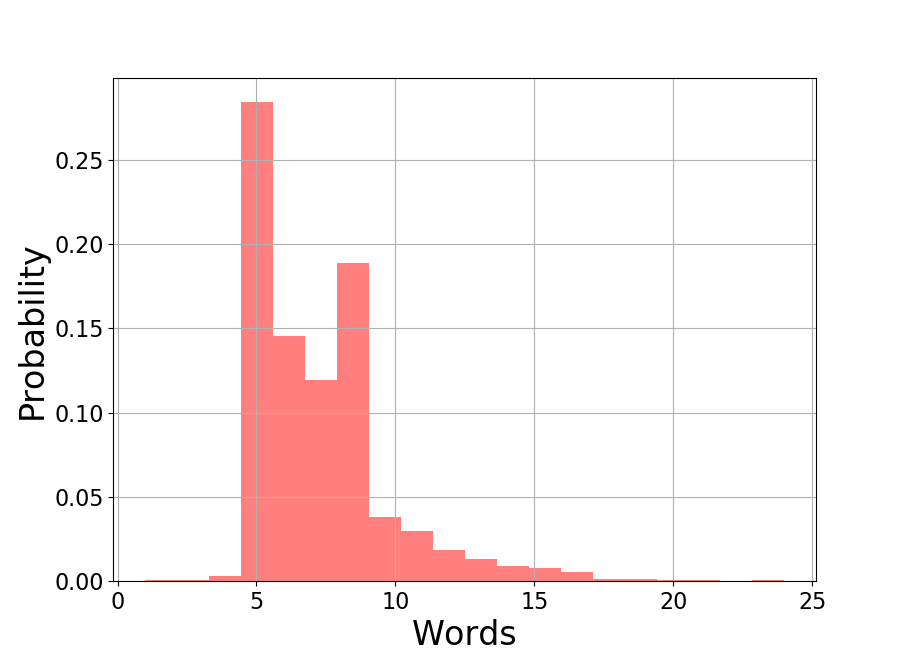} \\
\text{\quad  (a) Landmark descriptions \quad  \quad  \quad  (b) Local directional instructions  }\\
\includegraphics[width=0.8\linewidth,trim={.01\textwidth} {.00\textwidth} {0.05\textwidth} {.03\textwidth},clip]{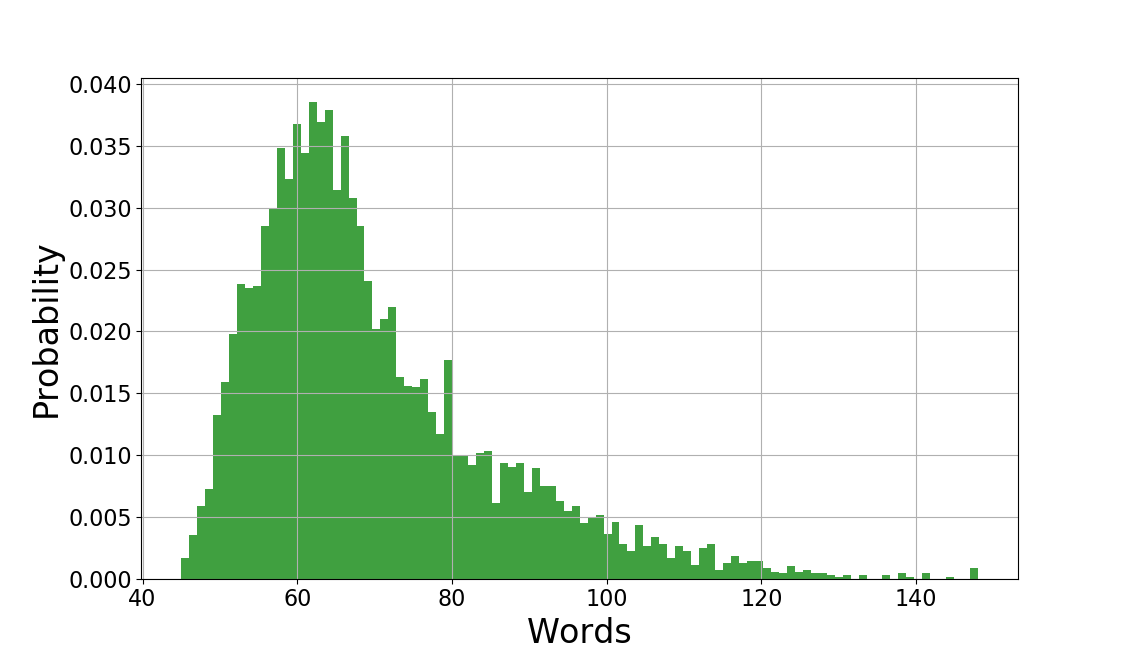}
\text{ \quad (c) Complete navigational instructions}
\caption{The length distribution of landmark instructions (a),  local directional instructions (b), and complete navigational instructions (c).}
\label{fig:instruction}
\end{figure}

\bigskip
\noindent
\textbf{Statistics}. We gathered $43,630$ locations which include GPS coordinates (latitudes and longitudes), bearing angles, etc. in New York City (NYC) covering an area of $10$ km $\times$ $10$ km as shown in Figure~\ref{fig:dataset_structure}. Out of all those locations, we managed to compile  $21,233$ street-view images (outdoor) -- each for one road node. We constructed a city graph with $43,630$ nodes and $80,000$ edges. The average \# of navigable directions per node is $3.4$.

We annotated $10,714$ navigational routes. These route descriptions are composed of $34,930$ node descriptions and $27,944$ local directional descriptions.  
Each navigational instruction comprises of 5 landmark descriptions (we use only 4 in our work) and 4 directional instructions. These 5 landmark descriptions include the description about the starting road node, the destination node and the three intermediate landmarks. Since the agent starts the VLN task from the starting node, we use only 4 landmark descriptions along with 4 directional instructions. The starting point description can be used for automatic localization which is left for future work. 
The average length of the navigational instructions, the landmark descriptions and the local directional instructions of the sub-routes are $68.8$ words, $8$ words and $7.2$ words, respectively. In total, our dataset Talk2Nav contains $5,240$ unique words. Figure~\ref{fig:instruction} shows the distribution of the length (number of words) of the landmark descriptions, the local directional instructions and the complete navigational instructions. 


In Table~\ref{tab:stat}, we show a  detailed comparison with the R2R dataset~\cite{anderson2018vision} and the TouchDown dataset~\cite{chen2018touchdown} under different criteria. While the three datasets share similarities in multiple aspects, the R2R and TouchDown datasets only annotate one overall language description for the entire route. It is costly to scale the annotation method to longer routes and to a large number of unknown environments. Furthermore, the annotations in these two datasets lacks the correspondence between language descriptions and sub-units of a route. Our dataset offers it without increasing the annotation effort. The detailed correspondence of the sub-units facilitates the learning task and enables an interesting cross-difficulty evaluation as presented in Section \ref{sec:ablation:study}.    


The average time taken for landmark ranking task is 5 seconds and for the route description task is around 3 minutes. In terms of payment, we paid \$$0.01$ and \$$0.5$ for the landmark ranking task and the route description task, respectively.  Hence, the hourly rate of payment for annotation is \$10. We have a qualification test. Only the annotators who passed the qualification test were employed for the real annotation. In total, $150$ qualified workers were employed. During the whole course of annotation process, we regularly sample a subset of the annotations from each worker and check their annotated route descriptions manually. If the annotation is of low quality, we reject the particular jobs and give feedback to the workers for improvement.  

We restrict the natural language to \emph{English} in our work. However, there are still biases in the collected annotations as is generally studied in the works of Bender \textit{et al.}~\cite{bender2011achieving,bender2018data}.   
The language variety in our route descriptions comprise of social dialects from United States, India, Canada, United Kingdom, Australia and others, showing diverse annotator demographics. Annotators are instructed to follow guidelines for descriptions like i) Do not describe based on street names or road names, ii) Try to use non-stationary objects mainly as visual landmarks. We also later perform certain curation techniques to the provided descriptions such as a) removing all non-unicode characters, b) correcting misspelled words using opensourced LanguageTool~\cite{languagetool}. 



\begin{table}[tb]
\caption{A comparison of our Talk2Nav dataset with the R2R dataset~\cite{anderson2018vision} and the TouchDown dataset~\cite{chen2018touchdown}. Avg stands for Average, desc for description, and (w) for (words).} 
\label{tab:stat}
  \centering
  \setlength\tabcolsep{0.8pt}
  \footnotesize
  \begin{tabular}{lccc}
\toprule
 Criteria & R2R~\cite{anderson2018vision} & Touchdown~\cite{chen2018touchdown} &Talk2Nav  \\
\midrule
\# of navigational routes & 7,189 & 9,326 &10,714 \\
\# of panoramic views & 10,800 & 29,641 &21,233 \\
\# of panoramic views per path & 6.0 & 35.2 & 40.0 \\
\# of navigational descriptions & 21,567 & 9,326 &10,714 \\
\# of navigational desc per path  & 3 &1 &1 \\
Length of navigational desc (w) & 29 & 89.6 & 68.8 \\
\# of landmark desc & - & - &34,930 \\
Length of landmark desc (w) & - & -&8 \\
\# of local directional desc & - &-&27,944 \\
Length of landmark desc (w) & - & - &7.2 \\
Vocabulary size & 3,156 & 4,999 &5,240 \\
Sub-unit correspondence &No & No & Yes\\
\bottomrule
\end{tabular} \vspace{0mm}
\end{table} 



\section{Approach}
\label{sec:approach}
\smallskip
\noindent

Our system is a single agent traveling in an environment represented as a directed connected graph. 
We create a street-view environment on top of the compiled city graph of our Talk2Nav dataset. The city graph consists of $21,233$ nodes/locations with the corresponding street-view images (outdoor) and $80,000$ edges representing the travelable road segments between the nodes. Nodes are points on the road belonging to the annotated region of NYC. Each node contains a) a 360\degree street-view panoramic image, b) the GPS coordinates of the corresponding location (defined by latitudes and longitudes), c) and the bearing angles of this node in the road network. A valid route is a sequence of connected edges from a source node to a destination node. Please see Figure~\ref{fig:dataset_structure} for the visualization of these concepts.  

During the training and testing stages, we use a simulator to navigate the agent in the street view environment in the city graph. Based on the predicted action at each node, the agent moves to the next node in the environment. A successful navigation is defined as when the agent correctly follows the right route referred by the natural language instruction. For the sake of tractability, we assume no uncertainty (such as dynamic changes in the environment, noise in the actuators) in the environment, hence it is a deterministic goal setting.

Our underlying simulator is the same as the ones used by previous methods~\cite{anderson2018vision,chen2018touchdown,learning:to:follow:19}. It, however, has a few different features. The simulator of \cite{anderson2018vision} is compiled for indoor navigation. The simulators of~\cite{chen2018touchdown,learning:to:follow:19} are developed for outdoor (city) navigation but they differ in the region used for annotation. The action spaces are also different. 
The simulators of~\cite{anderson2018vision,chen2018touchdown,learning:to:follow:19} use an action space consisting of \textit{left}, \textit{right}, \textit{up}, \textit{down}, \textit{forward} and \textit{stop} actions. We employ an action space of nine actions: eight moving actions in the eight equi-spaced angular directions between 0\degree and 360\degree and the \emph{stop} action. For each moving action, we select the road which has the minimum angular distance to the predicted moving direction. In the following sections, we present the method in details. 


\subsection{Route Finding Task}
The task defines the goal of an embodied agent to navigate from a source node to a destination node based on a navigational instruction. Specifically, given a natural language instruction $X = \{x_{1}, x_{2},...,x_{n}\}$, the agent needs to perform a sequence of actions $\{a_{1}, a_{2},..., a_{m}\}$ from the action space $\mathcal{A}$ to hop over nodes in the environment space to reach the destination node. Here $x_{i}$ represents individual word in an instruction while $a_{i}$ denotes an element from the set of nine actions 
(Detailed in Section~\ref{sec:action}). When the agent takes an action, it interacts with the environment and receives a new visual observation. The agent performs the actions sequentially until it reaches the destination node successfully or it fails the task because it exceeds the maximum episode length. 

The agent learns to predict an action at each state to navigate in the environment. Learning long-range vision-and-language navigation (VLN) requires an accurate sequential matching between the navigation instruction and the route. As argued in the introduction, we observe that navigation instructions consist of two major classes: landmark descriptions and local directional instructions between consecutive landmarks. In this work, given a language instruction, our method segments it to a sequence of two interleaving classes: landmark descriptions and local directional instructions. Please see Figure~\ref{fig:teaser} for an example of a segmented navigational instruction. 
As it moves, the agent learns an associated reference position in the language instruction to obtain a softly-attended local directional instruction and landmark description. When one landmark is achieved, the agent updates its attention and moves towards the new goal, i.e. the next landmark. Two matching modules are used to score each state by matching: 1) the traversed path in memory and the local, softly-attended directional instruction, and 2) the visual scene the agent observes at the current node and the local, softly-attended landmark description. An explicit external memory is used to store the traversed path from the latest visited landmark to the current position. 

\begin{figure*}[t] 
\centering
\includegraphics[width=0.9\linewidth,trim={.00\textwidth} {.0\textwidth} {0.0\textwidth} {.00\textwidth},clip]{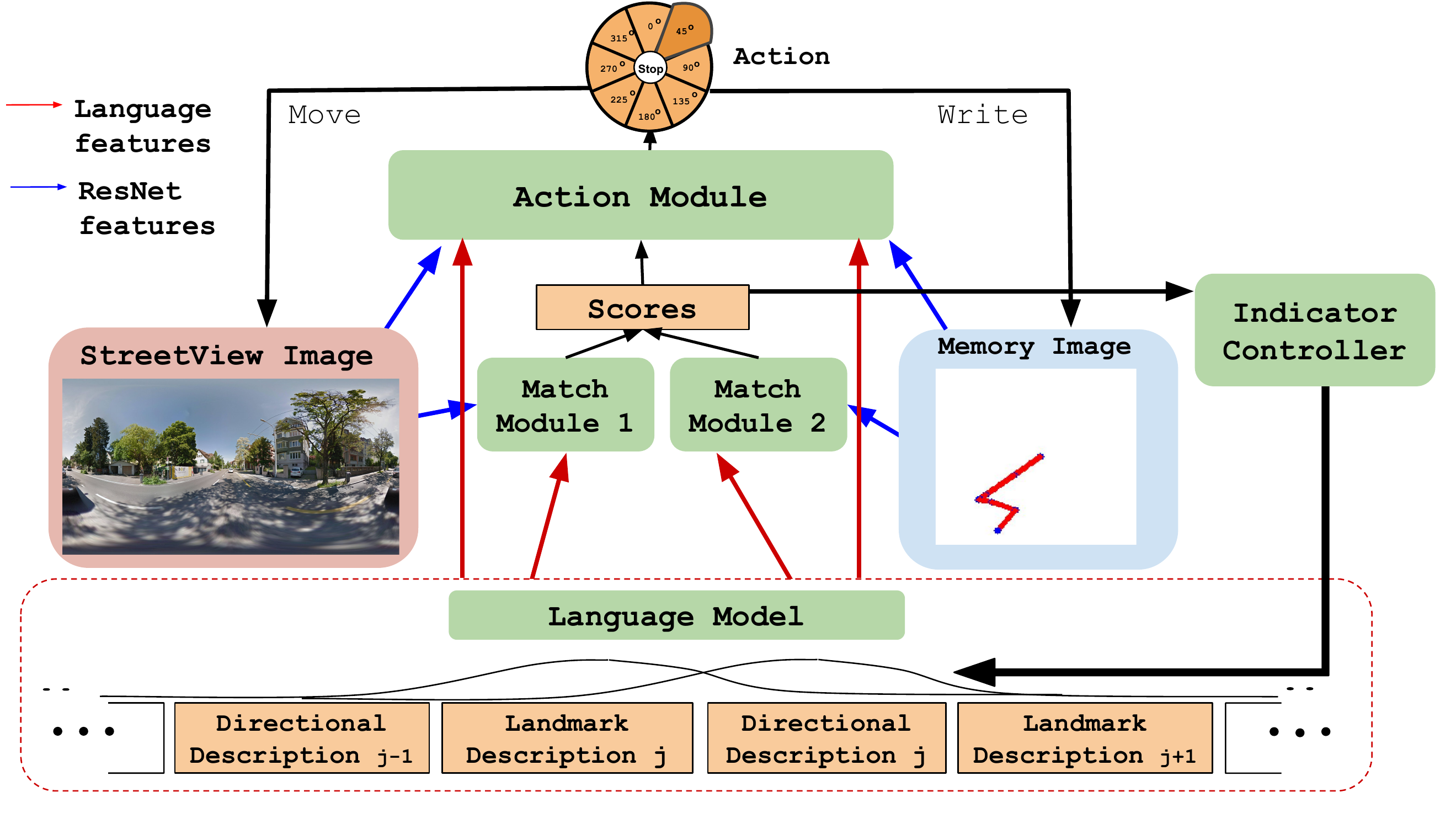} 
\caption{The illustrative diagram of our model. We have a soft attention mechanism over segmented language instructions, which is controlled by the Indicator Controller. The segmented landmark description and local directional instruction are matched with the visual image and the memory image respectively by two matching modules. The action module fetches features from the visual observation, the memory image, the features of the language segments, and the two matching scores to predict an action at each step. The agent then moves to the next node, updates the visual observation and the memory image, continues the movement, and so on until it reaches the destination. }
\label{fig:complete_model}
\end{figure*}

Each time a visual observation is successfully matched to a landmark, a learned controller increments the reference position of the soft attention map over the language instruction to update both the landmark and directional descriptions. This process continues until the agent reaches the destination node or the episodic length of the navigation is exceeded. A schematic diagram of the method is shown in Figure \ref{fig:complete_model}. Below we detail all the models used by our method. 

\smallskip

\subsubsection{Language}
\label{sec:language}
We segment the given instruction to two classes: a) visual landmark descriptions and b) local directional instructions between the landmarks. We employ the BERT transformer model~\cite{devlin2018bert} to classify a given language instruction $X$ into a sequence of token classes $\{c_{i}\}_{i=1}^{n}$ where $c_{i} \in \{0,1\}$, with $0$ denoting the landmark descriptions and and $1$ the local directional instructions. By grouping consecutive segments of the same token class, the whole instruction is segmented into interleaving segments. An example of the segmentation is shown in Figure~\ref{fig:teaser} and multiple more in Figure~\ref{fig:instruction_segments}. Those segments are used as the basic units of our attention scheme rather than the individual words used by previous methods~\cite{anderson2018vision}. This segmentation converts the language description into a more structured representation, aiming to facilitate the language-vision matching problem and the language-trajectory matching problem. Denoted by $T(X)$ a sequence of segments of landmark and local directional instructions, then 
\begin{equation}
\label{eq:langseg}
T(X) = \left( (L^{1},D^{1}), (L^{2},D^{2}), ..., (L^{J},D^{J}) \right)
\end{equation}
\noindent
where $L^{j}$ denotes the feature representation for segment $j$ of landmark description,  $D^{j}$ denotes the feature representation for segment $j$ of directional instruction, and $J$ is the total number of segments in the route description.


As it moves in the environment, the agent is associated with a reference position  in  the  language  description in order to put the focus on the most relevant landmark descriptions and the most relevant local directional instructions. This is modeled by a differentiable soft attention map. Let us denote by $\eta_{t}$ the reference position for time step $t$. The relevant landmark description at time step $t$ is extracted as  
\begin{equation}
\label{eqn:attn1}
\bar{L}^{\eta_t} = \sum_{j=1}^J L^je^{-|\eta_{t}-j|}
\end{equation}
and the relevant directional instruction as 
\begin{equation}
\label{eqn:attn2}
\bar{D}^{\eta_t} = \sum_{j=1}^J D^je^{-|\eta_{t}-j|},
\end{equation}
where $\eta_{t+1} = \eta_{t}+ \phi_t(.)$, $\eta_{0} = 1$, and $\phi_t(.)$ is an Indicator Controller learned to output $1$ when a landmark is reached and $0$ otherwise. The controller is shown in Figure~\ref{fig:complete_model} and is defined in Section \ref{sec:matching:module}. After a landmark is reached, $\eta_{t}$  increments $1$ and the attention map then centers around the next pair of landmark and directional instructions. The agent stops the navigation task successfully when $\phi_t()$ outputs 1 and there are no more pair of landmark and directional instructions left to continue. We initialize $\eta_{0}$ as $1$ to position the language attention around the first pair of landmark and directional instruction.




\subsubsection{Visual Observation}
The agent perceives the environment with an equipped $360\degree$ camera which obtains the visual observation $I_{t}$ of the environment at time step $t$. From the image, a feature $\psi_{1}(I_{t})$ is extracted and passed to the matching module as shown in Figure~\ref{fig:complete_model} which estimates the similarity (matching score) between the visual observation $I_{t}$ and a softly attended landmark description $\bar{L}^{\eta_{t}}$ which is modulated by our attention module. The visual feature is also passed to the action module to predict the action for the next step as shown in Section~\ref{sec:action}.

\subsubsection{Spatial Memory} 
\label{sec:spatialmemory}
Inspired by~\cite{gordon2018iqa,graves2016hybrid}, 
an external memory $M_{t}$ explicitly memorizes the agent's traversed path from the latest visited landmark. 
When the agent reaches a landmark, the memory $M_{t}$ is reinitialized to memorize the traversed path from the latest visited landmark. This reinitialization can be understood as a type of attention to focus on the recently traversed path in order to better localize and to better match against the relevant directional instructions $\bar{D}^{\eta_{t}}$ modeled by the learned language attention module defined in Section \ref{sec:language}.  

\begin{figure}[t] 
\includegraphics[width=1\linewidth,trim={.00\textwidth} {.00\textwidth} {0.00\textwidth} {.00\textwidth},clip]{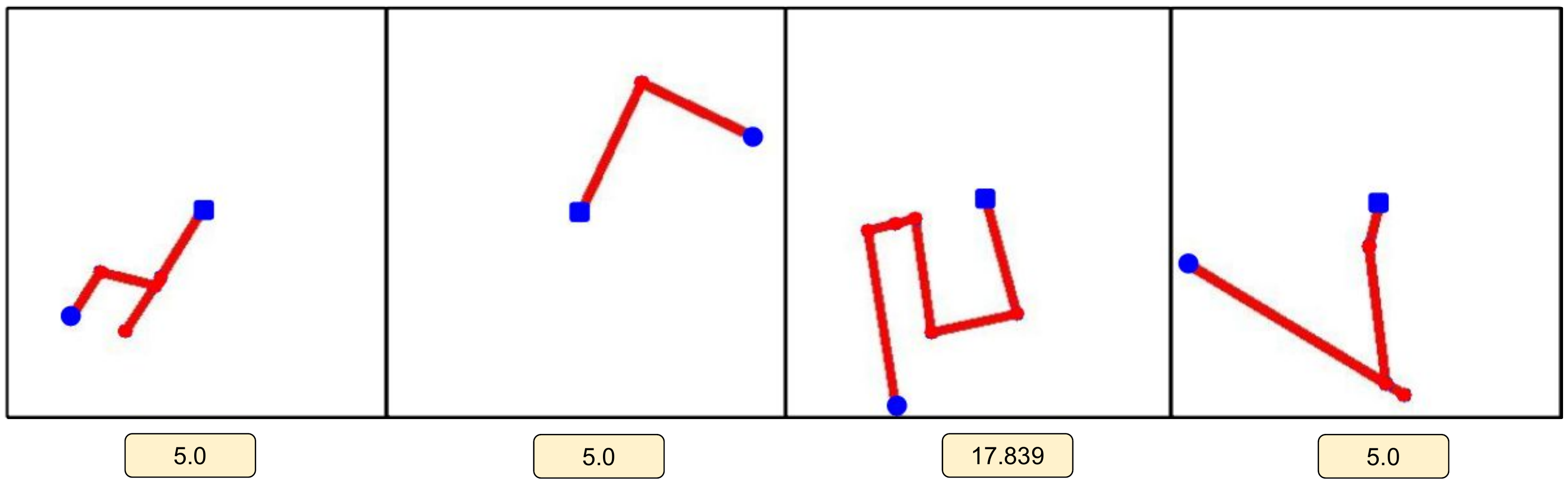} 
\caption{Exemplar memory images. The blue square denotes the source node, and the blue disk denotes the current location of the agent. The red lines indicate the traversed paths. Each memory image is associated with its map scale (metres per pixel) which is shown in the text box below.}
\label{fig:memory}
\end{figure}

As the agent navigates in the environment, we have a write module to write the traversed path to the memory image. For a real system navigating in the physical world, the traversed path can be obtained by an odometry system. In this work, we compute the traversed path directly from the road network in our simulated environment.  
Our write module traces the path travelled from the latest visited landmark to the current position. The path is rasterized and written into the memory image.  In the image, the path is indicated by red lines, the starting point by a blue square marker, and the current location is by a blue disk.  
The write module always writes from the centre of the memory image to make sure that there is room for all directions. Whenever the coordinates of the new rasterized pixel are beyond the image dimensions, the module incrementally increases the scale of the memory image until the new pixel is in the image and has a distance of $10$ pixels to the boundary.  An image of $200 \times 200$ pixels is used and the initial scale of the map is set to $5$ meters per pixel. Please find examples of the memory images in Figure~\ref{fig:memory}.

Each memory image is associated with the value of its scale (meters per pixel). Deep features $\psi_{2}(M_{t})$ are extracted from the memory image $M_{t}$ which are then concatenated with its scale value to form the representation of the memory. The concatenated features are passed to the matching module. The matching module verifies the semantic similarity between the traversed path and the provided local directional instruction. 
The concatenated features are also provided to the action module along with the local directional instruction features to predict the action for the next step. 

\bigskip
\noindent
\subsubsection{Matching Module}
\label{sec:matching:module}

Our matching module is used to determine whether the aimed landmark is reached. As shown in Figure \ref{fig:complete_model}, the matching score is determined by two complementary matching modules: 1) between the visual scene $\psi_{1}(I_{t})$ and the extracted landmark description $\bar{L}^{\eta_t}$ and 2)  between the spatial memory $\psi_{2}(M_{t})$ and the extracted directional instruction $\bar{D}^{\eta_t}$. 
For both cases, we use a generative image captioning model based on the transformer model proposed in~\cite{vaswani2017attention} and compute the probability  of reconstructing the language description given the image. Scores are the averaged generative probability over all words in the instruction. Let $s^{1}_{t}$ be the score for pair $(\psi_{1}(I_{t}), \bar{L}^{\eta_{t}})$ and $s^{2}_{t}$ be the score for pair $(\psi_{2}(M_{t}), \bar{D}^{\eta_{t}})$. We then compute the score feature $\mathbf{s}_{t}$ by concatenating the two:
$\mathbf{s}_{t} = (s^{1}_{t},s^{2}_{t})$. The score feature $\mathbf{s}_{t}$ is fed into a controller $\phi(.)$ to decide whether the aimed landmark is reached: 
\begin{equation}
\label{eqn:ctrl}
\phi_t(\mathbf{s}_{t}, \mathbf{h}_{t-1}) \in \{0,1\},
\end{equation}
\noindent
where $1$ indicates that the aimed landmark is reached and $0$ otherwise. $\phi_t(.)$ is an Adaptive Computation Time (ACT) 
LSTM \cite{graves2016adaptive} which allows the controller to learn to make decisions at variable time steps. $\mathbf{h}_{t-1}$ is the hidden state of the controller. In this work, $\phi_t(.)$ learns to identify the landmarks with the variable number of intermediate navigation steps.

\subsubsection{Action Module} 
\label{sec:action}
The action module takes the following inputs to decide the moving action: a) the pair of $(\psi_{1}(I_t),\bar{L}^{\eta_{t}})$, b) the pair of $(\psi_{2}(M_t),\bar{D}^{\eta_t})$, and c) the matching score feature $\mathbf{s}_{t}$. The illustrative diagram is shown in Figure~\ref{fig:complete_model}. 

The inputs in the a) group is used to predict $\mathbf{a}^{e}_{t}$ -- a probability vector of moving actions over the action space. The inputs in the b) is used to predict $\mathbf{a}^{m}_{t}$ -- the second probability vector of moving actions over the action space. The two probability vectors are adaptively averaged together, with weights learned from the score feature $\mathbf{s}_{t}$. Specifically, $\mathbf{s}_{t}$ is fed to a fully-connected (FC) network to output the weights $\mathbf{w}_{t} \in \mathbb{R}^{2}$.  For both  a) and b), we use an encoder LSTM as used in ~\cite{anderson2018vision} to encode the language segments. We then concatenate the encoder's hidden states with the image encodings (i.e. $\psi_1(I_t)$ and $\psi_{2}(M_{t})$) and pass through a FC network to predict the probability distribution $\mathbf{a}^{e}_{t}$ and  $\mathbf{a}^{m}_{t}$. By adaptively fusing the, we get the final action prediction $\mathbf{a}_{t}$:
\begin{equation}
\label{eqn:action}
\mathbf{a}_{t} = \frac{1}{\sum_{i} w^{i}_{t}}(w^{0}_{t}*\mathbf{a}^{e}_{t} + w^{1}_{t}*\mathbf{a}^{m}_{t}).
\end{equation}

The action $\mathbf{a}_{t}$ at time $t$ is defined as the weighted average of $\mathbf{a}^{e}_{t}$ and $\mathbf{a}^{m}_{t}$. For a different part of the trajectory, one of the actions ($\mathbf{a}_{t}^{e}$ or  $\mathbf{a}_{t}^{m}$) or both of them are reliable. This is heavily dependent on the situation. For instance, when the next landmark is not visible, the prediction should rely more on $\mathbf{a}_{t}^{m}$; when the landmark is clearly recognizable, the opposite holds. The learned matching scores will decide adaptively at each time step which prediction to be trusted and by how much. This adaptive fusion can be understood as a calibration system for the two complementary sub-systems for action prediction. The calibration method needs to be time- or situation-dependent.  A simpler summation/normalization of the two predictions is  rigid, and cannot distinguish between a confident true prediction and a confident false prediction. Confident false predictions are very common in deeply learned models.

The action space $\mathcal{A}$ is defined as follows. We divide the action space $\mathcal{A}$ into eight moving actions in eight directions and the stop action. Each direction is centred at $ \{(i*45\degree): i \in [0,...,7]\}$ 
with $\pm 22.5\degree$ offset as illustrated in Figure~\ref{fig:complete_model}. When an action angle is predicted, the agent turns to the road which has the minimum angular distance to the predicted angle  and moves forward to the next node along the road. The turning and moving-forward define an atomic action. The agent comes to a stop when it encounters the final landmark as stated in Section~\ref{sec:language}. 

It is worth noting that the implementation of our stop action is different from previous VLN methods~\cite{anderson2018vision,chen2018touchdown,learning:to:follow:19}.  As pointed out in the deep learning book \cite[p.384]{Goodfellow-et-al-2016}, the stop action can be achieved in various ways. One can add a special symbol corresponding to the end of a sequence, i.e. the STOP used by previous VLN methods. When the STOP is generated, the sampling process stops. Another option is to introduce an extra Bernoulli output to the model that represents the decision to either continue generation or halt generation at each time step. In our work, we adopt the second approach. The agent uses Indicator Controller (IC) to learn to determine whether it reaches landmarks. IC outputs $1$ when the agent reaches a landmark and updates the language attention for finding the next landmark until the destination is reached. The destination is the final landmark described in the instruction. The agent stops when the IC predicts $1$ and there is no more language description left to continue. Thus, our model has a stop mechanism but in a rather implicit manner.




\subsection{Learning}
The model  is trained in a supervised way. We followed the student-forcing approach proposed in~\cite{anderson2018vision} to train our model. At each step, the action module is trained with a supervisory signal of the action in the direction of the next landmark. This is in contrast with previous methods~\cite{anderson2018vision}, in which it is the direction to the final destination. 

We use the cross-entropy loss to train the action module and the matching module as they are formulated as classification tasks. 
For the ACT model, we use the weighted binary cross-entropy loss at every step. The supervision of the positive label (`1') for ACT model comes into effect
only if the agent reaches the landmark. The total loss is the sum of all module losses:
\begin{equation}
\label{eqn:loss}
Loss = 
Loss_{\text{action}} + Loss_{\text{matching}}^{\text{landmark}} + Loss_{\text{matching}}^{\text{direction}} + Loss_{\text{ACT}}.
\end{equation}
The losses of the two matching modules take effect only at the place of landmarks which are much sparser than the road nodes where the action loss and ACT loss are computed. Because of this, we first train the matching networks individually for the matching tasks, and then integrate them with other components for the overall training.   



\begin{figure}[t] 
\includegraphics[width=1\linewidth,trim={.00\textwidth} {.00\textwidth} {0.00\textwidth} {.00\textwidth},clip]{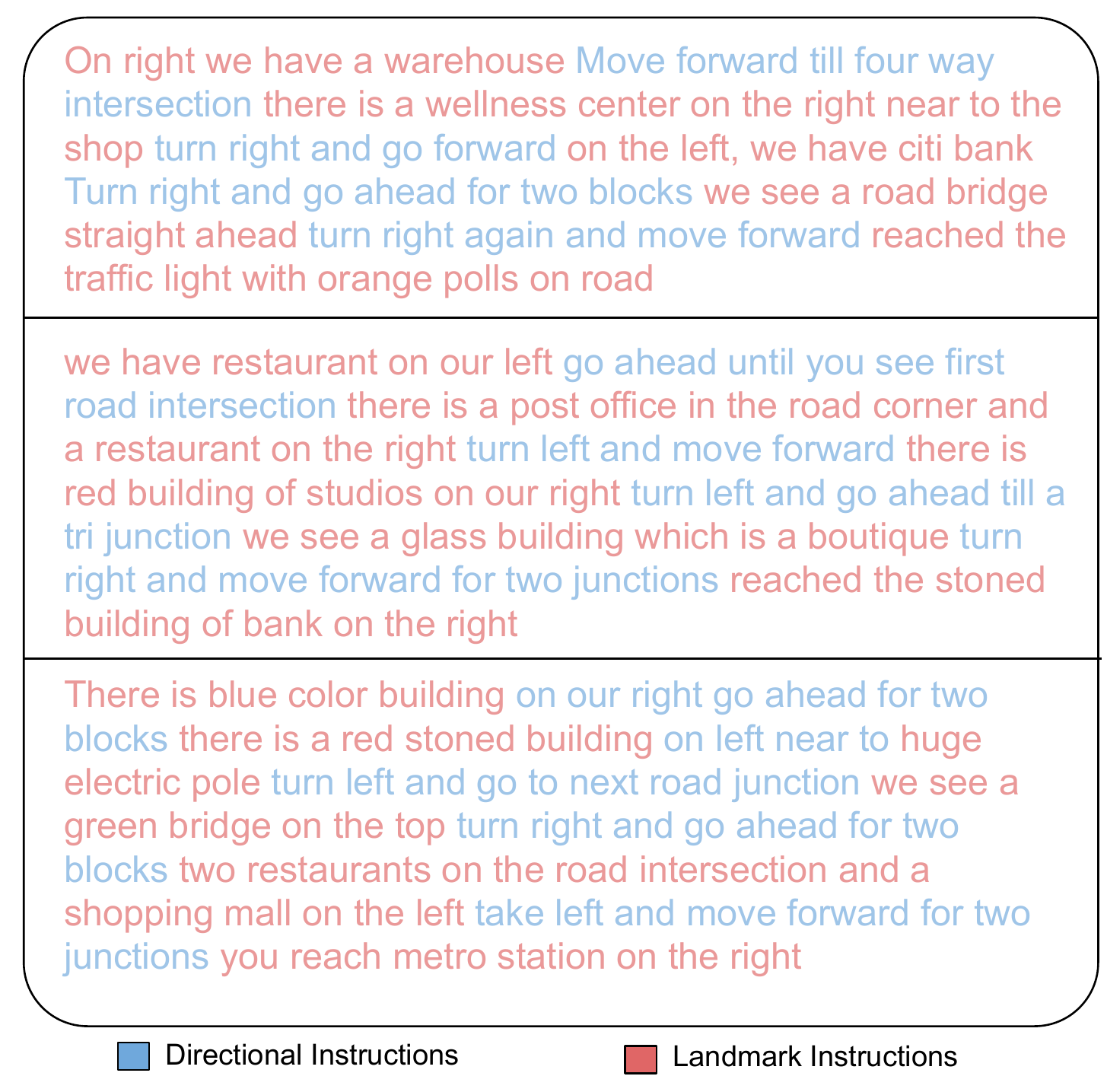} 
\caption{Examples of given instructions from Talk2Nav dataset. Different colours denote word token segmentation. Red denotes landmark descriptions and blue the local directional instructions.}
\label{fig:instruction_segments}
\end{figure}

\subsection{Implementation Details}

\smallskip
\noindent
\textbf{Language Module}. We use the BERT~\cite{devlin2018bert} transformer model pretrained on the BooksCorpus~\cite{zhu2015aligning} and the English Wikipedia\footnote{https://en.wikipedia.org/wiki/English\_Wikipedia}, for modelling language. This yields contextual word representations which is different from classical models such as word2vec~\cite{mikolov2013distributed} and GloVe~\cite{pennington2014glove}. We use a word-piece tokenizer to tokenize the sentence, by following~\cite{devlin2018bert,wu2016google}. Out of vocabulary words are split into sub-words based on the available vocabulary words. 
For word token classification, we first train BERT transformer with a token classification head. Here, we use the alignment between the given language instruction $X$ and their corresponding set of landmark and directional instruction segments in the training set of the Talk2Nav dataset. We then train the transformer model to classify each word token in the navigational instruction to be a landmark description or a local directional instruction. 

At the inference stage, the model predicts a binary label for each word token. We later convert the sequence of word token classes into segments $T(X)$ by simply grouping adjacent tokens which have the same class. 
We note that this model has a classification accuracy of $91.4\%$ on the test set. We have shown few word token segmentation results in Figure~\ref{fig:instruction_segments}. 

\bigskip
\noindent
\textbf{Visual inputs}. The Google street-view images are acquired in the form of equirectangular projection. We experimented with SphereNet \cite{coors2018spherenet} architecture pretrained on the MNIST \cite{lecun1998gradient} dataset 
and ResNet-101 \cite{he2016deep} pretrained on ImageNet~\cite{deng2009imagenet} to extract $\psi_{1}(I)$. Since the SphereNet pre-trained on MNIST is a relatively shallow network, we adapted the architecture of SphereNet to suit Street View images by adding more convolutional blocks. 
We then define a pretext task using the Street View images from the Talk2Nav dataset to learn the weights for SphereNet. 
Given two street-view images with an overlap of their visible view, the task is to predict the difference between their bearing angles and the projection of line joining the locations on the bearing angle of second location. We frame the problem as a regression task of predicting the  two angles.
This encourages SphereNet
to learn the semantics in the scene. We compiled the training set for this pre-training task from our own training split of the dataset. In the case of memory image, we use ResNet-101 
pretrained on ImageNet to extract $\psi_{2}(M)$ from the memory image $M$.

\bigskip
\noindent
\textbf{Other modules}. We perform image captioning in the matching modules using the transformer model, by following ~\cite{luo2018discriminability,zhu2018captioning,vaswani2017attention}. We pretrain the transformer model
for captioning in the matching module using the landmark street-view images and their corresponding descriptions from the training split of Talk2Nav for matching of the landmarks. For the other matching module of local directions, we pretrain the transformer model using the ground-truth memory images and their corresponding directional instructions. We synthesized the ground truth memory image in the same way as our write module writes to the memory image (as mentioned in Section~\ref{sec:spatialmemory}). We finetune both the matching modules in the training stage.
All the other models such as Indicator Controller, Action module are trained in an end-to-end fashion.

\bigskip
\noindent
\textbf{Network details}. The SphereNet model consists of five blocks of convolution and max-pooling layers, followed by a fully-connected layer. We use $32, 64, 128, 256, 128$ filters in the 1 to 5 convolutional layers and each layer is followed by a Max pooling and ReLU activation. This is the backbone structure of the SphereNet. For ResNet, we use ResNet-101 as the backbone. Later, we have two branches: the fully connected layer has $8$ neurons using a softmax activation function in one branch of the network for the action module and $512$ neurons with ReLU activation in the other branch for the match module. 
Hence, input of image feature size from these models are of $512$ and attention feature size over the image is $512$. The convolutional filter kernels are of size 5x5 and are applied with stride 1. Max pooling is performed with kernels of size 3x3 and a stride of 2. 
For language, we use an input encoding size of $512$ for each token in the vocabulary.  
We use Adam optimizer with learning rate of $0.01$ and alpha and beta for Adam as $0.999$ and $10^{-8}$. We trained for $20$ epochs with a mini-batch size of $16$.

\section{Experiments}
\label{sec:expts}

Our experiments focus on 1) the overall performance of our method when compared to the state-of-the-art (s-o-t-a) navigation algorithms, and 2) multiple ablation studies to further understand our method. The ablation studies cover a) the importance of using explicit external memory, b) a performance evaluation of navigation at different levels of difficulty, c) a comparison of visual features, d) a comparison of the performance with and without Indicator Controller (IC), and across  variants of matching modules.
We train and evaluate our model on the Talk2Nav dataset. We use $70\%$ of the dataset for training, $10\%$ for validation and the rest for testing. 
By following~\cite{anderson2018evaluation}, we evaluate our method  under three metrics: 
\begin{itemize}
    \item \textit{SPL}: the success rate weighted by normalized inverse path length. It penalizes the successes made with longer paths.
    \item \textit{Navigation Error}: the distance to the goal after finishing the episode. 
    \item \textit{Total Steps}: the total number of steps required to reach the goal successfully.
\end{itemize}

We compare with the following s-o-t-a navigation methods: Student Forcing ~\cite{anderson2018vision}, RPA~\cite{look:before:leap:eccv18}, Speaker-Follower~\cite{speaker:follower:nips18} and Self-Monitoring~\cite{ma2019self}. We trained their models on the same data that our model uses. The work~\cite{wang2019reinforced} yields very good results on the Room-to-Room dataset for indoor navigation. We, however, have not found publicly available code for the method and thus can only compare with other top-performing methods on our dataset.

We also study the performance of our complete model and other s-o-t-a methods at varying difficulty levels: from short navigation paths consisting of one landmark to long ones consisting of four landmarks.
In order to evaluate methods at different difficulty levels, we generate datasets of different navigation difficulties from the Talk2Nav dataset.
The navigation difficulty of a route is approximated by the length of the route which is measured by the number of landmarks it contains. In particular, in Talk2Nav, each route consists of 4 landmarks. For our primary experiments, we use the whole route for training and testing. For specific experiments to evaluate at different difficulty levels, we obtain routes with the length of 1, 2 and 3 landmarks by sub-sampling the annotated 4-landmark routes. For instance, we sample four 1-landmark sub-routes, three 2-landmark sub-routes and two 3-landmark sub-routes from a 4-landmark route.  We also generate a dataset with all four levels of navigation difficulty by mixing the original Talk2Nav and the three generated datasets. We use these sub-sampled routes to generate the datasets for the cross-difficulty evaluation. 


\subsection{Comparison to Prior Works}

To make a fair comparison with prior works, we use the same image features and language features in all the cases. We use pre-trained ResNet-101 model on ImageNet to extract image features and pre-trained BERT transformer model for language features. 
For Self-Monitoring~\cite{ma2019self} and Speaker-Follower~\cite{speaker:follower:nips18}, the panoramic view of the environment is discretized into 8 view-angles (8 headings x 1 elevation with 45 degree intervals). The navigable directions at each location are defined by the city graph of the Talk2Nav dataset. We use greedy action selection during evaluation as beam search decoding for action selection leads to lower SPL because of longer trajectory lengths~\cite{ma2019regretful}. 

Table~\ref{tab:expt} shows the results of our method and other competing s-o-t-a methods. We tabulate results under all the evaluation metrics. The destination threshold (when the distance of the final reached position and the destination node is within this range, it is a successful navigation) is set to $100$ m and
and the additional trajectory length (denoting additional allowed path length \emph{w.r.t} the ground-truth length to the destination) is set to $30\%$. 
This sets the budget of the trajectory length to be 130\% of the length of the ground truth. The  budget of the trajectory length is the maximum trajectory length allowed for the exploration episode.
We adopted the VLN evaluation criteria from ~\cite{anderson2018evaluation}. 
The row for \textit{Oracle} in Table~\ref{tab:expt} denotes the maximum accuracy that could be achieved when the ground-truth segments of the instructions are used instead of the segments by the trained segmentation method presented in Section~\ref{sec:language}. The row for \textit{Human} denotes the performance of human navigators. We asked the AMT workers to navigate from the starting node to the destination in our web-annotation interface with the annotated instructions. The performance is measured on the test set.

The table shows that our method achieves significantly better results than other existing methods under all considered evaluation metrics for long-range Vision-and-Language navigation in outdoor environments.  For instance, our method improves SPL from $9.56\%$ to $11.92\%$, reduces Navigation Error from $740.12m$ to $633.85m$ and reduces Total Steps from $43.1$ to $42.2$, when compared to the best competing method. 
We also conducted human studies on this task. We see that the \textit{Human} performance is much higher than all learning methods. This big gap means that the dataset offers a sufficient room for developing and improving learning algorithms on the dataset. This also validates the quality and the value of the dataset.

We also evaluate our method on the TouchDown dataset \cite{chen2018touchdown}. The Touchdown dataset has navigational instructions quite similar to the TalkNav dataset in terms of text length and the navigation environment. 
However, the Touchdown dataset has no correspondence between sub-descriptions and visual landmarks, and between sub-descriptions and sub-routes. Hence, 
we cannot train our method on the Touchdown dataset. We train our model on the Talk2Nav dataset and then evaluate it on the Touchdown dataset. We keep the model setting and the approach the same as used for our Talk2Nav dataset. We tabulate our evaluation results in Table~\ref{tab:expt:touchdown} under the same metrics. We compare the performance of our method with Student-Forcing~\cite{anderson2018vision}, RPA~\cite{look:before:leap:eccv18} and RConcat~\cite{chen2018touchdown}. We observe that our approach also outperforms other top-performing VLN methods on the TouchDown dataset~\cite{chen2018touchdown}.

In addition to the evaluation for the long-range (\emph{i.e.} 4-landmark) navigation, we also study the performance of all these trained methods when evaluated at varying difficulty levels: from short navigation paths consisting of one landmark to long ones consisting of four landmarks.  We evaluate under SPL and compare our method with the prior works as before.  
The results are listed in Table~\ref{tab:exptSPL}. We observe that our method outperforms all the other prior works by a large margin. For instance, our method improves SPL (\%) from $72.21$ to $74.28$ for short routes having 1 landmark, from $37.71$ to $43.08$ for routes with 2 landmarks, and from $23.16$ to $27.96$ for routes with 3 landmarks, when compared to~\cite{ma2019self}.

\begin{table}
\caption{Comparison with prior s-o-t-a methods.} 
\label{tab:expt}
  \centering
  \setlength\tabcolsep{2.5pt}
  \begin{tabular}{lccccccccccccc}
\toprule
 Methods &  SPL$\uparrow$ & Navigation Error$\downarrow$ &  Total Steps$\downarrow$ \\
\midrule
Random  & 2.88 & 1986.23 & 52.1\\
Student-forcing \cite{anderson2018vision} &  8.77 & 887.57 & 45.9\\
RPA \cite{look:before:leap:eccv18} & 8.43  & 803.90& 46.4\\
Speaker-follower \cite{speaker:follower:nips18} & 9.02 & 784.23& 43.9\\
Self-Monitoring \cite{ma2019self} & 9.56  & 740.12 & 43.1\\
Ours &  \textbf{11.92}  & \textbf{633.85} & \textbf{42.2} \\
\hline
Oracle & 13.12  & 520.90 & 40.6 \\
Human & 61.1 & 124.5 & 35.5 \\
\bottomrule
\end{tabular} 
\end{table} 

\begin{table}
\caption{Comparison with prior s-o-t-a methods. The numbers of RConcat is taken from the paper~\cite{chen2018touchdown}.} 
\label{tab:expt:touchdown}
  \centering
  \setlength\tabcolsep{2.5pt}
  \begin{tabular}{lccccccccccccc}
\toprule
 Methods &  SPL$\uparrow$ & Navigation Error$\downarrow$ &  Total Steps$\downarrow$ \\
\midrule
RConcat~\cite{chen2018touchdown} & 10.4 & 234.57 & 48.2 \\

\hline

Student-forcing \cite{anderson2018vision} &  11.61 & 207.86 & 45.1 \\
RPA \cite{look:before:leap:eccv18} & 13.27  & 187.45 & 44.7 \\
Ours &  \textbf{20.45}  & \textbf{102.41} & \textbf{41.9} \\
\bottomrule
\end{tabular} 
\end{table} 

\begin{table}
\caption{Comparison of our method to other methods at varying difficulty levels: from short paths consisting of one landmark to long paths consisting of four landmarks. SPL$\uparrow$ is used as the metric.} 
\label{tab:exptSPL}
  \centering
  \setlength\tabcolsep{3.5pt}
  \begin{tabular}{lccccccccccccc}
\toprule
 Methods / \#(Landmarks) & 1 & 2 & 3& 4 \\
\midrule
Random walk & 31.05 & 16.95 & 9.04 & 2.88\\
Student-forcing \cite{anderson2018vision} & 55.63 & 26.76 & 18.31 & 8.77\\
RPA \cite{look:before:leap:eccv18} &59.75 & 30.30 & 18.20 & 8.43\\
Speaker-follower \cite{speaker:follower:nips18} & 71.14 & 35.79 & 21.47 & 9.02 \\
Self-Monitoring \cite{ma2019self} & 72.21 & 37.71 & 23.16 & 9.56\\
Ours & \textbf{74.28} & \textbf{43.08}& \textbf{27.96} & \textbf{11.92} \\
\bottomrule
\end{tabular} 
\end{table}


\subsubsection{Analysis}

The reason behind the good performance of our method can be attributed to multiple factors. The decomposition of the whole navigation instruction into landmark descriptions and local directional instructions, the attention map defined on language segments instead of English words, and the two clearly purposed matching modules make our method suitable for long-range vision-and-language navigation. Due to these introduced components and the design that allows them to work together, the agent is able to put the focus to the right place and does not get lost easily as it moves.  

Previous methods aim to find a visual image match for the given language sentence at each step to predict an action. We argue that this is not optimal. The navigational instructions indeed consist of mixed referrals for visual content (landmarks) and for spatial movements. For instance, it is wrong to match a sentence like `Go forward 100m and take a left turn' to an image observation. Our method distinguishes the two types of language sentences and computes a better matching score between language description and the sequential actions (spatial movement + visual observations). In this work, we use an explicit memory, as explained in Section~\ref{sec:approach}, to keep track of the spatial movements from a topological perspective. 

It is possible that the comparison models would obtain similar improvements if additional supervision for landmark matching is given.
It is worth noting that it is not straightforward to incorporate the landmark grounding into other methods. More importantly, if one compare the total annotation cost, this `additional' supervision does not cost more annotation effort -- our annotation method is easier and more efficient than annotating the whole long-range route at once. 
Thus, being able to use this additional supervision without adding extra annotation cost is the  contribution of our work.

\subsection{Ablation Studies}
\label{sec:ablation:study}

In order to further understand our method, we perform three ablation studies on memory types, difficulty levels of navigation, with and without IC, variants of matching modules and visual features. 

\begin{figure}[!tb]
\includegraphics[width=1.0\linewidth,trim={.00\textwidth} {.00\textwidth} {0.00\textwidth} {.00\textwidth},clip]{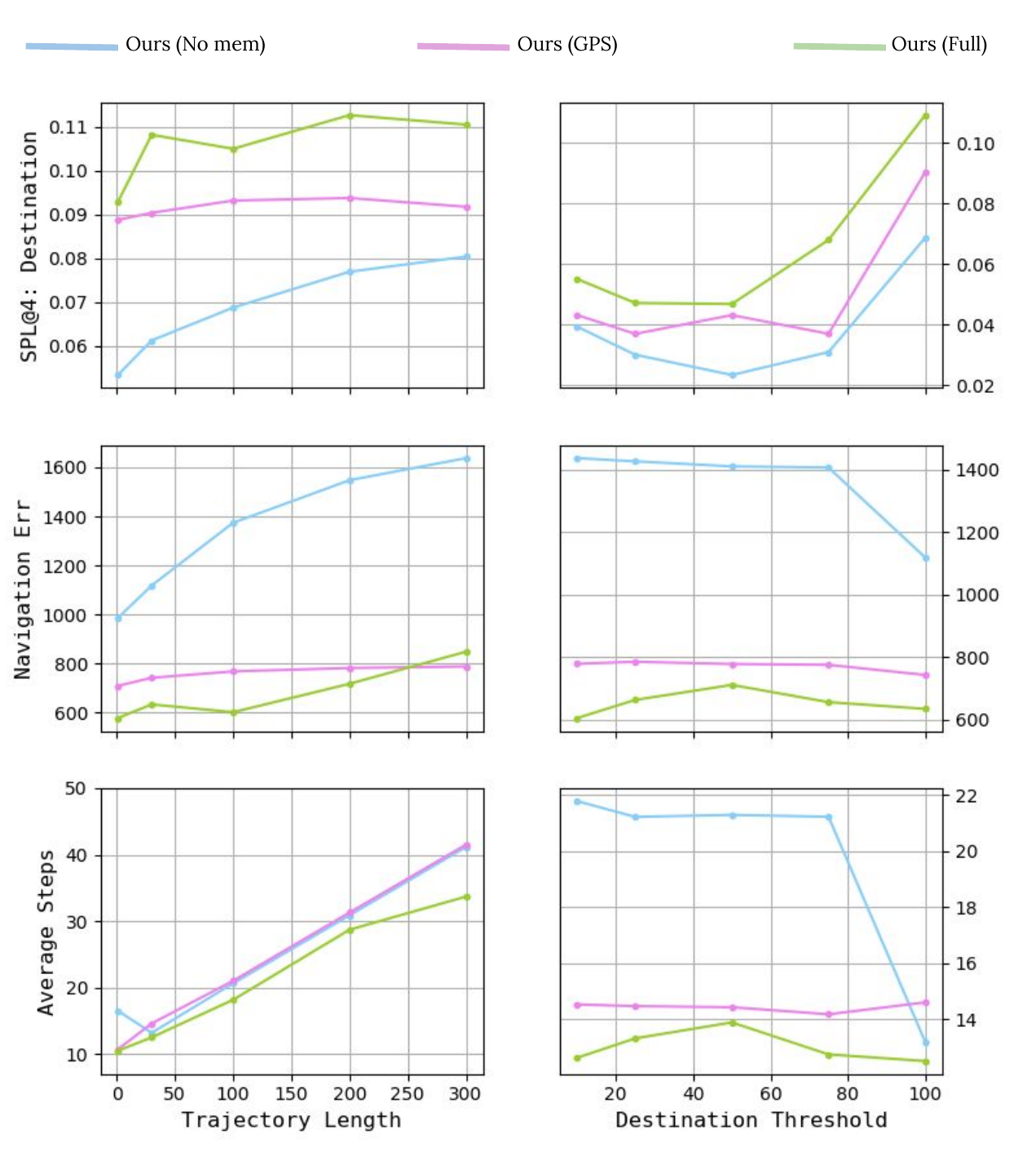} 
\caption{Rows for metrics and columns for evaluation settings. Trajectory length: \% extra length of path allowed for agents to navigate at evaluation. Destination Threshold (meter): a threshold of the final distance to the destination, below which is considered as a success.}
\label{fig:graph}
\end{figure}

\bigskip
\noindent
\textbf{Memory}. 
We compare \textit{our memory} to \textit{no memory} and to \textit{a trajectory of GPS coordinates}.  For GPS coordinates, an encoder LSTM is used to extract features $\psi_{3}(G_t)$, where $G_{t}=\{(X_{i}^{gps},Y_{i}^{gps}): i \in [1,...,t] \}$.
The results are reported in Table~\ref{tab:variants}.
We see that Student-Forcing~\cite{anderson2018vision} and RPA~\cite{look:before:leap:eccv18} has a SPL score of $8.77\%$ and $8.43\%$, respectively (from Table~\ref{tab:expt}). \textit{Ours} with \emph{no memory} gets $6.87\%$. This is because Anderson \textit{et al.}~\cite{anderson2018vision} and Wang \textit{et al.}~\cite{look:before:leap:eccv18} use sequence-to-sequence model~\cite{sutskever2014sequence} which has implicit memory about the history of visual path and actions. The poor performance of \textit{no memory} is also seen in Figure~\ref{fig:graph} when compared against all other methods either with implicit or explicit memory.

Encoding the explicit memory as a trajectory of GPS coordinates improves SPL from $6.87\%$ to $9.04\%$ as done in \emph{Ours (no memory)}. \textit{Ours} with GPS memory also performs better than having implicit memory under both SPL and Navigation Error as shown in Table~\ref{tab:variants}.
Furthermore, when we employ the top-view trajectory map representation as the explicit memory, our method outperforms all the previous methods either with no memory or with implicit memories as one can see in Table~\ref{tab:expt} and Table~\ref{tab:variants}.  
This validates the effectiveness of our explicit memory. 
It has been found already that top-view trajectory map representation is very useful in learning autonomous driving models~\cite{drive:surroundview:route:planner}. Our findings are in line with theirs, in a relevant but different application.

\begin{table}[!tb]
\caption{Comparison of our method under different settings: a) three variants of memory: no memory, GPS memory and our memory, and b) two different visual features: ResNet and SphereNet.} 
\label{tab:variants}
  \centering
  \setlength\tabcolsep{1.50pt}
  \footnotesize
  \begin{tabular}{ccccccccccccccccccc}
\toprule
 \multicolumn{3}{c}{Memory} &  \multicolumn{2}{c}{Visual Feature}  & \multirow{2}{*}{SPL$\uparrow$}  & \multirow{2}{*}{Nav Err$\downarrow$} & \multirow{2}{*}{Tot. Steps$\downarrow$} \\ 
 No & GPS & Ours & ResNet & SphereNet  &  &  & \\  \midrule
 \ding{51} & &  & \ding{51} & &   6.87 & 1374.02 & 48.74\\
& \ding{51}& & \ding{51}& &9.04 & 742.36 & 46.47\\
& & \ding{51}& & \ding{51} &9.37& 801.38 & 44.93\\
& & \ding{51}& \ding{51}& &\textbf{11.92} & \textbf{633.85} & \textbf{42.29}\\
\bottomrule
\end{tabular} 
\end{table}

\begin{table}
\caption{Cross-difficulty evaluation of our method. We train our model with training data of routes with each difficulty level $\{1,2,3,4\}$ and we evaluate for all the models on all difficulty levels. \emph{All} denotes the mixed routes of all difficulty levels.}
\label{tab:exptLandmarks}
\makegapedcells
\centering
\setlength\tabcolsep{5.5pt}
\begin{tabular}{@{}cc|ccccc@{}}
\multicolumn{1}{c}{} &\multicolumn{1}{c}{} &\multicolumn{5}{c}{\# of landmarks (test)} \\ 
\multicolumn{1}{c}{} & 
\multicolumn{1}{c|}{} & 
\multicolumn{1}{c}{1} & 
\multicolumn{1}{c}{2} &
\multicolumn{1}{c}{3} & 
\multicolumn{1}{c}{4} &
\multicolumn{1}{c}{All}\\ [1.0ex]
\cline{2-7}
\multirow[c]{5}{*}{\rotatebox[origin=tr]{90}{\# of landmarks (train)}}
& 1  & \textbf{76.71} & 34.21 & 21.41 & 8.45 & 45.89 \\[0.5ex]
& 2  & 75.11   & 46.42  & 27.07   & 10.71 & 50.08 \\ [0.5ex]
& 3  & 74.43 & \textbf{46.62}  & \textbf{28.14}  & 11.51 & 50.41  \\[0.5ex]
& 4   & 74.28   & 43.08  & 27.96   & \textbf{11.92} & 49.93\\ [0.5ex]
& All  & 75.33   & 46.10   & 27.32  & 11.68   &  \textbf{51.37} \\ [0.5ex]
\cline{2-7}
\end{tabular}
\end{table}
We take a step further to evaluate the memory types  under different evaluation settings of trajectory length and destination threshold. We again compare our method \textit{Ours (Full)} to \textit{Ours (No memory)} and \textit{Ours (GPS memory)}. The results are shown in Figure~\ref{fig:graph}. The figure shows that \textit{Ours (Full)} has higher SPL consistently over others at different trajectory length and destination threshold. Our variant \textit{Ours (GPS)} where GPS information is encoded as a sequence of coordinates, shows deteriorating performance with increasing trajectory length. Subsequently, this shows the importance of a 2-dimensional representation of the memory as used in \textit{Ours (Full)}. In the case of Navigation Error plots, we observe that \textit{Ours (Full)} performs better than other methods. However, we see that \textit{Ours (Full)} performs worse than \textit{Ours (GPS)} at 300\% of trajectory length. This may be because the memory module draws a long traversed path in the external memory which may lead to overlap of the line. This overlap reduces the matching accuracy and thus increases the navigation error. Finally, in the case of Average Steps, \textit{Ours (Full)} has lower value (better performance) compared to other methods when tested under different settings for the trajectory length and destination threshold. We observe that average steps of \textit{Ours (Full)} increases relatively slower than other methods with the trajectory length. It can be seen that \textit{Ours (Full)} outperforms other methods consistently over varying destination thresholds.

\bigskip
\noindent
\textbf{Navigation at different difficulty levels}. 
To further understand how our method behaves when training and testing on different navigation difficulty levels, we take one step further to study the effect of training and testing our model with routes of variable number of landmarks. We have tabulated in Table~\ref{tab:exptLandmarks} the performance of the method under the different combinations of route lengths (i.e. the number of landmarks) used in the training and testing phases. 

The table shows that the performance of our model is fairly high when trained with routes of $1$ landmark and tested on the same difficulty level. However, the performance drops drastically when tested on routes of higher difficulty levels. The performance on longer routes improves when we train the model with routes of the same or higher difficulty levels. The main conclusion from this experiment is that a model trained with harder cases works well on easier test cases, but not the other way around.

In the last row, we show the model performance when trained with mixed routes of all the difficulty levels. The trained model achieves competitive performance at all navigation difficulty levels and outperforms all other models for the mixed difficulty levels. 

One can also see that the performance drops almost exponentially with the level of navigation difficulty. The navigation at difficulty level $4$ is already very challenging as the highest SPL is $11.92$ only. Hence, annotating even longer routes is not very necessary at the moment for training and validating the current learning algorithms. This experiment is also a showcase of the notable merit of our Talk2Nav dataset that routes of different difficulty levels can be created for the fine-grained evaluation. 


\bigskip
\noindent
\textbf{With and without IC.}
Here, we explore the significance of Indicator Controller in our model. IC has two roles in our approach, the first being modelling attention over the segments of landmark and local directional instructions, and the other being an indicator of reaching the landmarks. We train and test our models with and without IC. We also experiment with different IC models such as LSTM and ACT (final model). 
In Table~\ref{tab:expt:ablation:IC}, \textit{w/o IC} represents our approach without IC. To adapt this setting, we combine all landmark instruction segments into one single instruction and all directional segments into another single instruction. We employ transformer model with self-attention for feature extraction from the landmark and directional instructions. Further to indicate reaching the landmarks, we also add an additional STOP action in action space. We observe a drop in the performance to $9.71$. This can be due to the implicit attention model over the landmark and directional instructions. For \textit{IC:LSTM} model, we replace the ACT with the regular LSTM~\cite{hochreiter1997long} for our IC module. We see that the performance drops to $8.95$. This is probably because the inherent merit of ACT over the regular LSTM that it is able to dynamically predict the number of recurrent steps based on the input and the hidden states.


\begin{figure*}[t]
  \centering
  \includegraphics[width=0.97\textwidth]{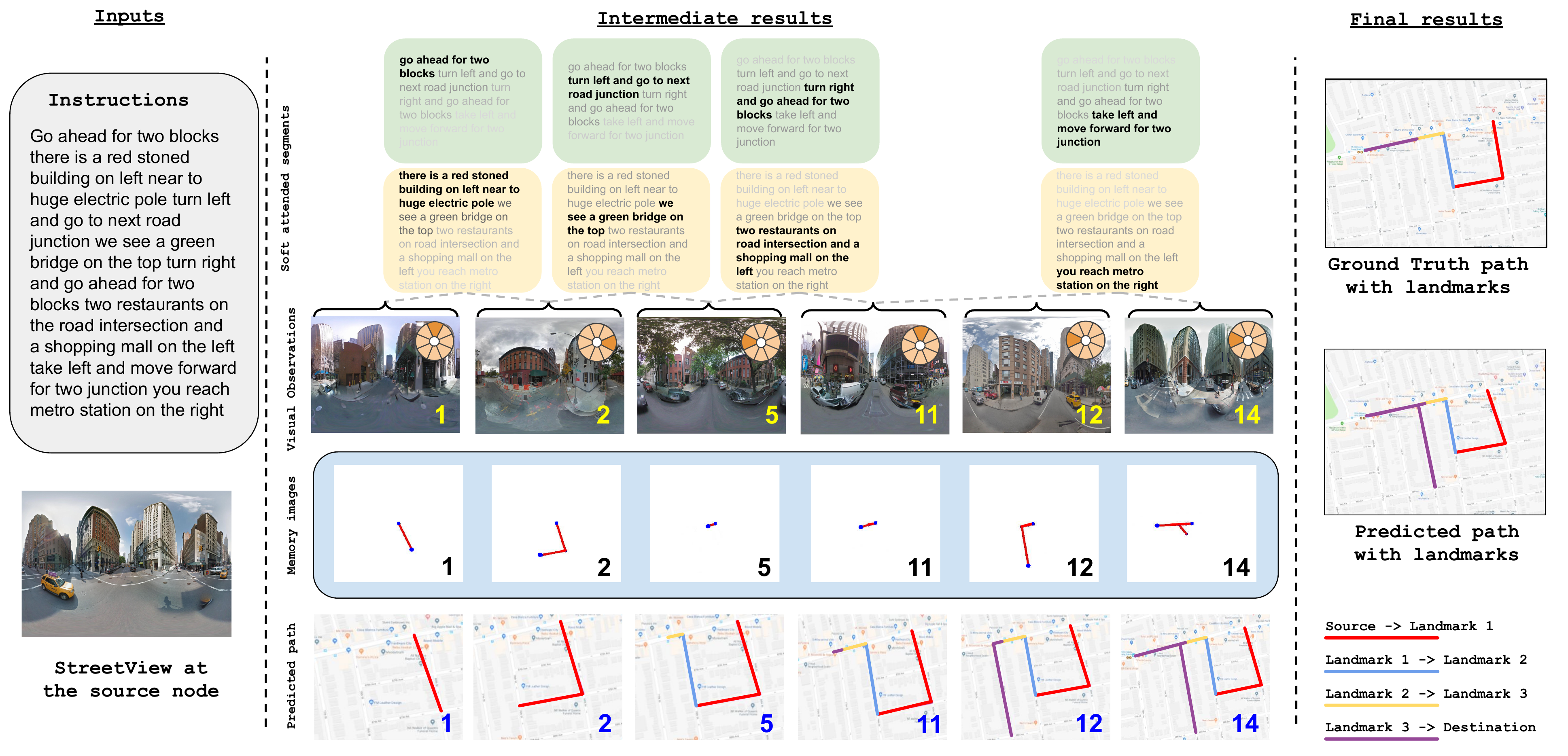} \\
  \includegraphics[width=0.97\textwidth]{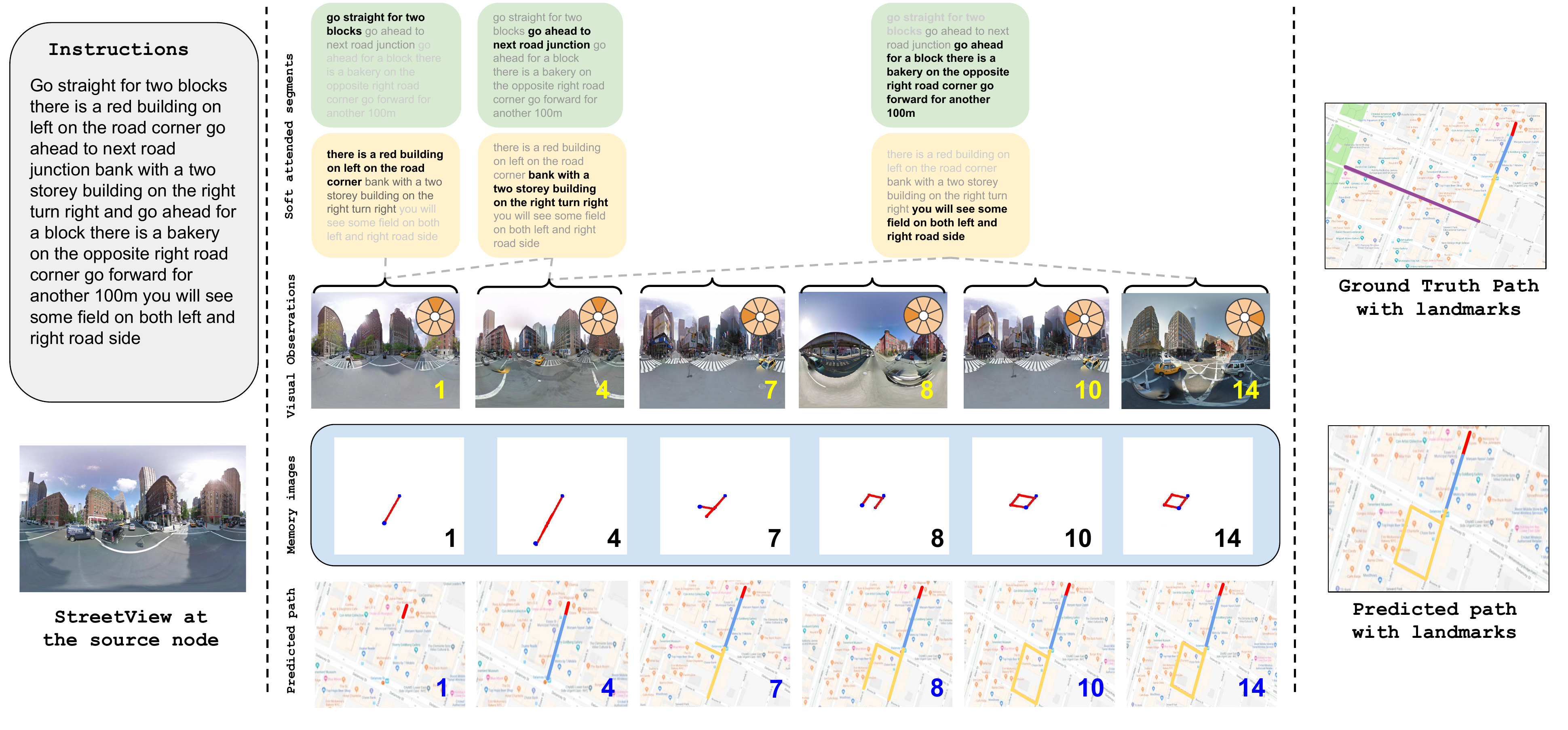} \\
\vspace{-0.2cm}
  \caption{Two qualitative results from our approach. The left panel shows the inputs: the navigational instruction and the 360\degree image at the source node. The middle panel shows the rows of intermediate results along the navigated path; from top to bottom, it shows the language segments of directional (green) and landmark instructions (yellow), the 360\degree views of the agent along the path with imposed action predictions, the memory images and the agent's navigated paths, of which different stages are indicated by different colors. The numbers embedded in the figure indicate the number of the step. We cannot not show all intermediate steps due to space constraints. The right panel shows the final traversed path by the agent and the ground-truth trajectory with different colours indicating different sub-routes between the consecutive landmarks. Note that the overhead maps are shown only for illustration purpose and they are not available to the agent.}
  \label{fig:qual-results}
\end{figure*}

\begin{table}[!tb]
\caption{Comparison of our method under different settings: a) three variants of IC: w/o IC, LSTM and ACT, and b) two different Matching modules: Discriminative (Disc) and Generative (Gen).} 
\label{tab:expt:ablation:IC}
  \centering
  \setlength\tabcolsep{1.50pt}
  \footnotesize
  \begin{tabular}{ccccccccccccccccccc}
\toprule
 \multicolumn{3}{c}{IC} &  \multicolumn{2}{c}{Match module}  & \multirow{2}{*}{SPL$\uparrow$}  \\ 
 w/o & LSTM & ACT & Disc. & Gen.  &  \\  \midrule
 \ding{51} & &  & &\ding{51} &   9.71 \\
& \ding{51}& & &\ding{51} &8.95 \\
& & \ding{51}& \ding{51}& & 11.15 \\
& & \ding{51}& &\ding{51} &\textbf{11.92} \\
\bottomrule
\end{tabular} 
\end{table}

\bigskip
\noindent
\textbf{Matching module variants.} We experiment with two variants for matching module of Section~\ref{sec:matching:module} here. We use a discriminative model and a generative image captioning model. Both the models receive features from the language and images and are supposed to compute scores based on their semantic matching. In the discriminative model, features from images and language are concatenated before given to multilayer perceptrons which outputs a matching score. The triplet ranking loss is used to learn the model. To sample negative samples, we use randomly-chosen street-view images and randomly-chosen memory images. We compare this discriminative model to the generative model used in our method. 
In Table~\ref{tab:expt:ablation:IC}, we observe that the generative model performs better than the discriminative model by 0.77 points. The main merit behind using the generative model is that it requires just the positives, hence is not sensitive to the choices of the negative sampling methods. A more carefully designed negative mining step may make the discriminative model more competitive. We leave this  as the future work.

\bigskip
\noindent
\textbf{SphereNet vs. ResNet}. We observe that using the pre-trained SphereNet yields better accuracy for the navigation task than training from scratch. This means that our proposed self-learning tasks are useful for model pre-training. However, we see in Table~\ref{tab:variants} that learning with SphereNet has comparable or slightly worse performance than with ResNet. This may be due to the fact that ResNet is trained on ImageNet which comes with human labels. 
Henceforth, in this work we use ResNet for the evaluation. We believe that SphereNet is likely to perform better with architectural changes like residual connections, which can be a promising future work.

\subsection{Qualitative Analysis}

We also provide two qualitative examples in a step-by-step fashion in Figure~\ref{fig:qual-results}. The given input navigational instructions and the 360\degree visual observations at the source node are given on the left of the figure. The rest is devoted to the intermediate and final results during the course of the VLN task. The middle panel of the figure shows the rows of intermediate results: a) the first two rows show the results of language segments of landmark descriptions and directional instructions, b) the third row depicts the front 360\degree views of the agent along the path embedded with the prediction of moving direction, c) the fourth row shows the memory images written by the write module and, d) the final row shows the agent's navigated paths predicted by our model. Different stages of finding landmarks are indicated by different colors. The ground truth route with landmarks are on the right of the figure. The numbers on the images denote the number of steps already traversed. We see that when the agent reaches a landmark, the memory is re-initialized and the soft attention over the navigational instruction segments moves forward to the next segment. Due to the space limit, we do not show all the intermediate steps.

In Figure~\ref{fig:qual-results}, we show a successful navigation in the first example. Although the agent misses the right path at some point, it successfully comes back to the path towards the destination. It takes 14 steps to reach the destination of which only a sub-set of intermediate steps are displayed in the figure. In the second case, we observe that the agent fails to navigate to the destination and forms a loop in the road until it finishes the episode. The failure is partially due to the segmentation error of the language instruction -- the overall instruction is segmented into $3$ segments instead of $4$. This causes confusion for the matching modules. In both of the cases, we can clearly see that the memory of the topological view of the path is intuitive to interpret.

We also present a video\footnote{\url{https://www.trace.ethz.ch/publications/2019/talk2nav/index.html}} to demonstrate our task and approach with more examples.

\section{Conclusion}
This work has proposed a new approach for language guided automatic wayfinding in cities using spatial memory framework and soft dual attention mechanism over language descriptions.
The main contribution of our work are: a) an effective method to create large-scale navigational instructions over long-range city environments; b) a new dataset with verbal instructions for $10,714$ navigational routes; c)  a novel learning approach of integrating explicit memory framework of remembering the traversed path, and soft attention model over the language segments controlled by Adaptive Computation Time LSTMs. Our method connects two lines of research that have been less explored together so far: mental formalization of verbal navigational instructions \cite{pictorial:verbal:tools:99,when:why:visual:landmarks:01,structural:salience:landmarks:route:directions:05} and training neural network agent for automatic wayfindings \cite{anderson2018vision,chen2018touchdown}. Experiments show that our model outperforms other methods consistently.  

\vspace{2mm}
\noindent
\textbf{Acknowledgement}: This work is funded by Toyota Motor Europe via the research project TRACE-Zurich.

\bibliographystyle{spmpsci}      
\bibliography{egbib}   

%
%

\end{document}